\documentclass[acmlarge,screen]{acmart}

\usepackage{graphicx}
\usepackage{subcaption}
\usepackage{multirow}
\usepackage{makecell}
\usepackage[capitalise]{cleveref}
\usepackage{adjustbox}
\usepackage{siunitx}
\usepackage{enumitem}

\AtBeginDocument{%
  }

\setcopyright{acmlicensed}
\copyrightyear{2024}
\acmYear{2024}
\acmDOI{XXXXXXX.XXXXXXX}

\acmJournal{IMWUT}
\acmVolume{0}
\acmNumber{0}
\acmArticle{0}
\acmMonth{0}

\begin{document}

\title[Beyond Confusion]{Beyond Confusion: A Fine-grained Dialectical Examination of Human Activity Recognition Benchmark Datasets}


\author{Daniel Geißler}
\email{daniel.geissler@dfki.de}
\orcid{1234-5678-9012}
\affiliation{%
  \institution{German Research Center for Artificial Intelligence (DFKI)}
  \city{Kaiserslautern}
  \country{Germany}
}

\author{Dominique Nshimyimana}
\email{dominique.nshimyimana@dfki.de}
\affiliation{%
  \institution{University of Kaiserslautern-Landau (RPTU)}
  \city{Kaiserslautern}
  \country{Germany}
}

\author{Vitor Fortes Rey}
\email{vitor.fortes_rey@dfki.de}
\affiliation{%
  \institution{German Research Center for Artificial Intelligence (DFKI),}
  \institution{University of Kaiserslautern-Landau (RPTU)}
  \city{Kaiserslautern}
  \country{Germany}
}

\author{Sungho Suh}
\email{sungho.suh@dfki.de}
\affiliation{%
  \institution{German Research Center for Artificial Intelligence (DFKI),}
  \institution{University of Kaiserslautern-Landau (RPTU)}
  \city{Kaiserslautern}
  \country{Germany}
}

\author{Bo Zhou}
\email{bo.zhou@dfki.de}
\affiliation{%
  \institution{German Research Center for Artificial Intelligence (DFKI),}
  \institution{University of Kaiserslautern-Landau (RPTU)}
  \city{Kaiserslautern}
  \country{Germany}
}

\author{Paul Lukowicz}
\email{paul.lukowicz@dfki.de}
\affiliation{%
  \institution{German Research Center for Artificial Intelligence (DFKI),}
  \institution{University of Kaiserslautern-Landau (RPTU)}
  \city{Kaiserslautern}
  \country{Germany}
}

\renewcommand{\shortauthors}{Geißler et al.}

\begin{abstract}
  The research of machine learning (ML) algorithms for human activity recognition (HAR) has made significant progress with publicly available datasets.
However, most research prioritizes statistical metrics over examining negative sample details.
While recent models like transformers have been applied to HAR datasets with limited success from the benchmark metrics, their counterparts have effectively solved problems on similar levels with near 100\% accuracy. 
This raises questions about the limitations of current approaches. 
This paper aims to address these open questions by conducting a fine-grained inspection of six popular HAR benchmark datasets. 
We identified for some parts of the data, none of the six chosen state-of-the-art ML methods can correctly classify, denoted as the intersect of false classifications (IFC).
Analysis of the IFC reveals several underlying problems, including ambiguous annotations, irregularities during recording execution, and misaligned transition periods. 
We contribute to the field by quantifying and characterizing annotated data ambiguities, providing a trinary categorization mask for dataset patching, and stressing potential improvements for future data collections. 
\end{abstract}




\maketitle

\section{Introduction}

A significant amount of research on machine learning (ML) methods for sensor-based human activity recognition (HAR) has emerged over the past decade with the growing availability of public datasets, especially based on inertial measurement units (IMUs).
The datasets such as PAMAP2 \cite{reiss2012pamap2} and Opportunity \cite{chavarriaga2013opportunity} have accelerated machine learning (ML) development as ML research, especially deep learning, can focus on the model architectures and be independent from the time-consuming data collection processes \cite{wang2019deep, bian2022state}.
They also allow a fair comparison across different proposed machine learning models as benchmarks.
However, most such ML-focused works only conclude by reporting the statistical level metrics and confusion matrix and fall short of examining the details, especially the negative samples.
However, in typical studies evaluating the viability of novel sensing modalities, such negative samples are crucial for the discussion.

\begin{figure}[!t]
    \centering
    \includegraphics[width=\textwidth]{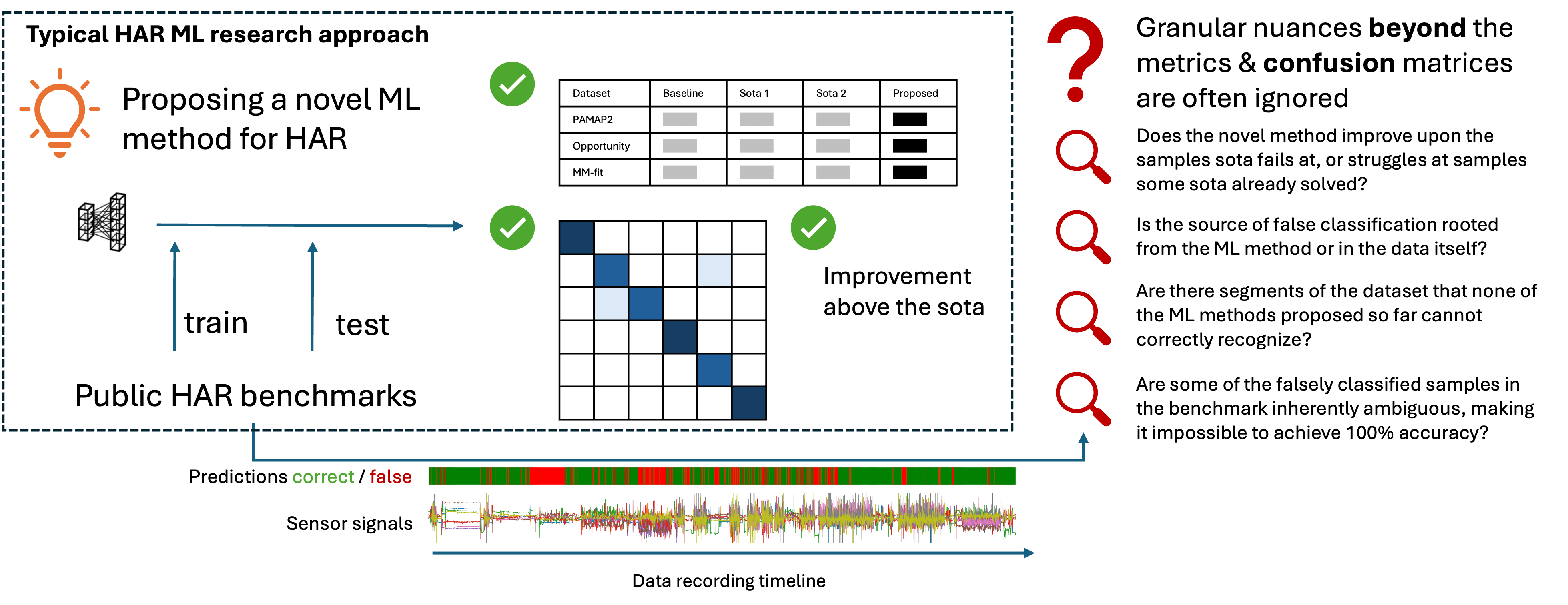}
    \caption{We aim to examine HAR research beyond the typical approach focusing on the statistical metrics and confusion matrices tested on public benchmark datasets to demonstrate improvement. }
    \Description{}
    \label{fig:intro}
\end{figure}

Some recent ML models proposed for the HAR datasets are adapted from architectures that have achieved high levels of recognition in datasets from other disciplines with similar classification tasks.
For example, for text classification, transformer models like DeBERTa and C-BERT \cite{karl2022transformers} have reached over 98\% accuracy in the 8-class subset of the Reuters-21578 dataset (R8) \footnote{http://www.daviddlewis.com/resources/testcollections/reuters21578/};
vision transformers like ViT-H/14 \cite{dosovitskiy2021an} have reached over 99\% accuracy in the popular 10-class image classification dataset CIFAR-10 \cite{krizhevsky2009learning}.
Although there are other benchmarks that have inferior reported results, like the ImageNet dataset \cite{deng2009imagenet} currently topping at 92\% by OmniVec \cite{srivastava2024omnivec}, there are 1000 classes of objects which is inherently more challenging than 10-class problems as the statistical chance level is significantly lower.
Yet, the adapted versions of these models like TinyHAR \cite{zhou2022tinyhar} which is a transformer-based model still struggle in many HAR classification benchmarks with similar or fewer numbers of classes (e.g. PAMAP2 at the range of 70 to 80\%).
This is especially true when proper leave-out partitions between the training, validation, and testing data were performed in combination with cross-validation to avoid overfitting.
The nature of different modalities can partially account for this difference, as wearable sensor signals are obscure while vision or text are easily interpretable by human perceptions, requiring advanced mechanism to enhance the data and model insight \cite{geissler2023latent}.
Nonetheless, the obscure nature of sensor signals essentially leaves researchers little device for verifying the data context but to have full faith in the provided annotations as ground truth.
Yet human annotators inevitably make mistakes or produce ambiguities as pointed out by Hoelzemann, et al. \cite{hoelzemann2024evaluation}
As shown in \cref{fig:intro},
combining the lack of attention given to negative samples, inferior performance from better-established modalities, and human factors in annotation, we raise the research question: `Are some of the falsely classified samples in the benchmark datasets inherently ambiguous, making it impossible to achieve 100\% accuracy?'

There have been discussions in this direction in the literature, including developing remedies.
Ward, et al. \cite{ward2011performance} pointed out the influence of granular discrepancies between the prediction and annotations on the short time window level, and proposed event-based metrics.
Kwon et al. \cite{kwon2019handling} attempted to address the annotation uncertainty problem by artificially inserting a Gaussian distribution as a smoothing kernel to each consecutive sequence of the same class.
A recent study by Hoelzemann, et al. \cite{hoelzemann2024evaluation} revealed the human factors in HAR dataset annotation.
Despite these works attempting to overcome the inherent annotation ambiguities in HAR, there has been no systematic analysis that quantifies the extent, characterizes the underlying problems, and provides viable solutions for existing datasets, which is the main motivation of our work.
To answer such questions and provide solutions to help move HAR forward, we conducted a fine-grained inspection of the datasets in this work.

This paper first provides a literature overview of the popular HAR benchmarks and how they have helped accelerate machine learning development in this field in the past two decades.
Then we use the most widely used ML methods which can be considered milestones in the related work (five model architectures as well as pre-training enhancement) to examine the datasets, especially the samples that cannot be successfully detected by any of the models.
We look beyond the summary of statistical metrics and inspect the dataset on the granularity of seconds in short windows.
We have found that there is a significant overlap of the false classifications among all of the models in all of the datasets.
We then closely examine those data windows that none of the models can correctly classify, and attempt to characterize the underlying problem.
Detailed examination of these common false classifications reveals several common phenomena:
\begin{enumerate}
    \item Ambiguous annotations: in a single-output classification task, the different classes should be mutually exclusive. However, we have found some classes can be composed of other classes, or different classes have common elements of motion.
    \item Execution during recording: in some classes that are supposed to be static, the participant may have short windows of transient spontaneous random motions. This in combination with annotation of large granularity (i.e. only annotating the start and end of long segments of activities) usually results in irregularities that are incoherent with the designated labels.
    \item Transition periods: Depending on the labeling design granularity, transitions between activity classes are commonly falsely classified if the transition period is excessively long and inconsistently labeled. 
\end{enumerate}

In the wake of these identified and characterized problems in the six datasets, we provide the following contributions that can be utilized by the community in further research:
\begin{itemize}
    \item Quantify the label ambiguity through the Intersect of False Classification (IFC) to provide a realistic accuracy target, effectively pinpointing the segments of each dataset that still cannot be classified by any of the ML methods.
    \item Upon analysis of the IFC, we propose a trinary label mask for different types of challenges that can be used as patching of the corresponding datasets, to categorize and filter sections from datasets into clean, minor and major.
\end{itemize}

\begin{table}[ht]
\footnotesize
\centering
\caption{A none-exhaustive chronicle of  HAR datasets in the past two decades}
\label{tab:related}
\footnotesize
\begin{tabular}{p{3cm} p{3.2cm} p{2cm} p{0.9cm}p{5cm}} 
\toprule
Dataset (Year) & Scope           & Classes (Num)              & Subjects & Sensor Placement       \\ \midrule
Skoda (2008)    \cite{zappi2008activity}    & car maintenance                    & gestures (10)   & 1        & 20 accelerometer on two arms                                                       \\
DarmstadtDR (2008) \cite{huynh2008discovery} & daily routines without scripts            &activities (34)&1&2 accelerometers at right hip and right wrist\\
HCI gestures (2009) \cite{forster2009unsupervised}& freehand or blackboard interaction & gestures (5)    & 1        & 8 accelerometers on one arm                                                       \\
\textbf{WISDM} (2011)       \cite{kwapisz2011wisdm} & daily locomotion activities        & locomotion (6)  & 29       & 1 smartphone accelerometer in the pocket                                           \\
HASC (2011)    \cite{hasc2011}     & daily locomotion activities        & locomotion (6)  & 540      &                                                                                   \\
\textbf{PAMAP2} (2012)    \cite{reiss2012pamap2}   & daily complex activities           & activities (18) & 9        & 3 IMUs on the chest, dominant wrist, and dominant ankle and one heart rate sensor
\\ 
RealDisp (2012) \cite{banos2012benchmark}     & daily whole-body movements and body part-specific actions & activities (33) & 17       & 9 IMUs across the body     \\                                 Daphnet (2012)\cite{durka2012daphne}& Gait recording of Parkinson's disease patients with occasional freeze & walk, freeze (2)                            & 10       & 3 accelerometers on legs and hip                                                  \\
\textbf{Opportunity} (2013) \cite{chavarriaga2013opportunity}   & daily complex activities                                  & locomotion (5) gestures (18) & 4        & 7 IMUs + 12 accelerometers                                                        \\
UCI-HAR (2013)  \cite{anguita2013public}      & daily locomotion activities                               & locomotion (6)                              & 30       & 1 waist-attached smartphone accelerometer and gyroscope                           \\
\textbf{MHealth} (2014)   \cite{banos2014mhealth}     & daily and sports activities                               & activities (12)                             & 10       & 3 IMUs on the chest, left ankle, right lower arm, with 2-lead ECG                 \\
USC-HAD (2015)\cite{jiang2015human}& daily locomotion and routine activities                                & activities (12)                             & 14       & 1 wrist-worn IMU                                                                  \\
DSADS (2015)    \cite{zhang2015recognizing}      & daily and sports activities                               & activities (19)                             & 8        & 5 IMUs across the body                                                            \\
UTD-MHAD (2015)    \cite{chen2015utd}   & locomotion interaction and sports activities              & actions (27)                                & 8        & 1 wrist or thigh IMU, Kinect depth camera                                         \\
RealWorld (2016)   \cite{sztyler2016body}   & daily locomotion activities                               & locomotion (7)                              & 15       & 7 IMUs across the body                                                            \\
ActiveMiles (2016)\cite{ravi2016deep}&daily locomotion and mobility activities & locomotion (7)                              & 10       & 1 smartphone IMU                                                                  \\
Mobiact (2016)\cite{vavoulas2016mobiact}& daily locomotion and living activities                                 & activities (9)                              & 50       & 1 smartphone IMU                                                                  \\
SBHAR (2016)\cite{reyes2016transition}& daily locomotion activities                                            & locomotion (6)                              & 30       & 1 waist-worn smartphone IMU                                                       \\
\textbf{MotionSense} (2018)\cite{malekzadeh2018motionsense}& daily locomotion activities                                            & locomotion (6)                              & 24       & 1 smartphone in the pocket                                                        \\
SHL (2018)\cite{gjoreski2018university}& daily locomotion and mobility activities                               & locomotion (8)                              & 3        & 4 smartphone IMUs, body-worn camera                                               \\
UK-Biobank (2018)\cite{yuan2024self,willetts2018statistical}& daily activities                                                       & not publicly available                      & 100000  & 1 accelerometer on wrist                                                         \\
\textbf{MM-Fit} (2020)\cite{stromback2020mmfit}& sports activities                                                      & activities (11)                             & 10       & 5 IMUs (earbud, smartphones, smartwatches), full body poses                       \\
Capture-24 (2021)\cite{chan2021capture,capture24}& daily activities (partially labelled)                                  & activities(33)                              & 124      & 1 accelerometer on wrist                                                          \\

\bottomrule
\end{tabular}
\end{table}

\section{Related Work}

\subsection{IMU-based HAR benchmark datasets}

First, we examine a non-exhaustive chronicle of HAR datasets in the past two decades in \cref{tab:related}.
The early works typically used a 3-axis accelerometer providing the local acceleration of the sensing node; while works after 2010 largely leveraged 6-axis inertial measurement units (IMUs) containing accelerometer and gyroscope, or 9-axis IMUs with the inclusion of magnetometer, which can be converted to acceleration and orientation in the global reference frame.
Some works also combine other modalities like heart rate or electrocardiogram (ECG), body temperature, and so on.
Most benchmark datasets are in the scope of daily activities, especially modes of locomotion (e.g. standing, walking, running, climbing stairs, etc.), while some are specific for certain domains like fitness and machine maintenance.

Some datasets were performed by participants following scripted instructions, and the annotations were the start and end time of large duration of the activities from the instructions like in \textit{PAMAP2};
while in some datasets the participants were allowed to perform activities freely under a defined scope, like parts of \textit{Opportunity}, and the annotations were performed usually with the help of synchronized videos on a fine time granularity.


\subsection{State-of-the-art classification methods for HAR}
Across the landscape of potential Artificial Intelligence applications, Human Activity Recognition (HAR) is a widely researched field commonly aiming to classify human activities, especially through body-worn and even more commonly design and everyday-usable sensors \cite{geissler2023moca,zhou2023mocapose}. Those are typically characterized by full or partial body movement constituting activities of daily living, fitness exercises, or locomotion. While HAR can be achieved using diverse modalities such as video \cite{ke2013review}, bio-impedance \cite{liu2024imove}, or capacitance \cite{geissler2024embedding}, in this work, we focus on Inertial Measurement Units (IMUs), as they are the most popular wearable sensing modality given its prevalence in wearable devices such as phones, watches, among others. Body-worn IMUs with precise orientation and acceleration measurements provide indispensable time-series information on the wearers' body posture and movements. However, many factors hamper overall IMU-based HAR performance. First, the information provided in a realistic everyday deployment, such as a single smartphone and smartwatch, provides less informative motion information when compared to a full body posture obtained by another modality such as video. Second, IMU-based HAR has a lack of  {\bf labeled} sensor data from complex realistic scenarios. The amount of data present for IMU data pales in comparison to that available for video where one data set, the Kinetics-700 \cite{smaira2020short} has 650,000 video clips covering 700 different human activities.  Third, given the natural behavior of human beings and the inherited physiological complexity of human bodies, different people perform the same activities in different ways. This leads to a gap in performance when evaluating unseen subjects instead of new data for subjects in the training set.

Numerous researchers have studied IMU sensor-based HAR. Generally, IMU sensor-based HAR can be categorized into classical statistical methods and data-driven deep learning-based approaches. The classical statistical methods focus on hand-crafted feature extracting using signal processing techniques and conventional machine learning models to capture activity data distributions, commonly utilizing time-domain features, such as mean, variance standard deviation, Kurtosis, and Skewness, and frequency-domain features like power spectral density and peak frequency \cite{lara2012survey}. Bulling et al. \cite{bulling2014tutorial} introduced a process framework treating individual sensor data frames as statistically independent for activity recognition, also exploring temporal models such as hidden Markov models. Hidden Markov models were also compared to Gaussian mixture models and random forest models by Attal et al. \cite{attal2015physical} with raw data and selected features. Plötz et al. \cite{plotz2011feature} presented a deep Boltzmann machine-based method to learn features from data automatically for HAR. Anguita et al. \cite{anguita2012human} proposed a multi-class SVM model on smartphones for locomotion activity recognition, while Hammerla et al. \cite{hammerla2013preserving} introduced an empirical cumulative density function (ECDF) feature to retain signal spatial information.  

Recently, due to the fast development and advancement of deep learning techniques, deep learning-based HAR methods have improved the HAR performance. Yang et al. \cite{yang2015deep} proposed a convolutional neural network (CNN)-based HAR method designed with multiple convolutional and pooling filters along temporal dimensions for sensor data processing. Recurrent deep learning methods, such as Long Short-Term Memory (LSTM) units \cite{hochreiter1997long}, have also been explored \cite{nakano2017effect}, with models like DeepConvLSTM \cite{ordonez2016deep} combining LSTM and CNN layers for temporal correlation capture. 
Other non-mainstream methods have also been adapted to HAR, such as graph neural networks \cite{wieland2023tinygraphhar}, spike neural networks \cite{bian2023evaluating} and liquid neural networks \cite{hasani2022closed}.
Hybrid approaches, such as employing deep belief networks in hidden Markov models \cite{alsheikh2016deep}, have shown promise. Moreover, ensemble methods \cite{guan2017ensembles} and attention mechanisms \cite{vaswani2017attention, zeng2018understanding} have been applied, with Mahmud et al. \cite{mahmud2020human} achieving improvements on previous methods using self-attention for HAR. 
Several works also explored blending attention with convolution and recurrent layers have shown further improvement in HAR \cite{al2022multi, zhou2022tinyhar, gao2021danhar} and are considered the current state-of-the-art in terms of model architectures.
Since the possibility of running the algorithm on mobile devices is a major concern for wearable computing, works like TinyHAR \cite{zhou2022tinyhar} and PatchHAR \cite{wang2024patchhar} also emphasize resource efficiency together with recognition performance. 
Additionally, the whole development pipeline needs to be considered to decrease energy demand and model size to an acceptable minimum \cite{geissler2024power}.

Alongside developments in network architecture, other fundamental challenges in HAR started to be addressed through data generation and representation learning paradigms\cite{abedin2021attend}. The lack of training data for IMU-based was tackled by simulation approaches such as \cite{xiao2021deep, rey2019let, fortes2021translating, kwon2020imutube, kwon2021approaching}, which can generate IMU data from online repositories such as YouTube. While this simulated data has shown to be helpful \cite{kwon2021approaching}, it suffers from problems beyond limitations in the simulation itself. First, it constrains activities to those that can be easily filmed. Second, human labeling is still necessary, which is not ideal. On the other hand, representation learning approaches have attempted to improve the training procedure itself. For example, including adversarial learning for improving user-generalization \cite{bai2020adversarial, suh2022adversarial, suh2023tasked} or leveraging user labels through mutual attention and multi-task learning \cite{sheng2020weakly, chen2020metier}. Another approach for generating better representations is self-supervised learning (SSL). With the recent popularity of SSL approaches for language and vision, there has also been a move to a pretrain-then-fine-tune approach where pretext tasks are performed in large amounts of unlabeled data and then fine-tuned for classification in the target dataset. This has been explored for IMU-based HAR in many self-supervised approaches such as \cite{assessing_har, multitask_original, cpc:har, enhanced_cpc}. Those methods have proven to be beneficial in some scenarios, especially in cases where only a very small amount of labeled data is available. Recently, a multi-modality has also been a topic of research for HAR \cite{deldari2022cocoa}. In fact, IMU-based HAR can benefit from multi-modal training even when the extra modalities/sensors will not be present at test time. This has been achieved through methods such as \cite{fortes2022learning, lago2021using, yang2022more, liu2024imove}. This is achieved either through contrastive learning \cite{fortes2022learning, nguyen2023virtual, cheng2023learning, liu2024imove, wieland2023tinygraphhar}, cosine similarity \cite{yang2022more} or clustering \cite{lago2021using}.

While much has been achieved for HAR through more than two decades of research, less attention has been paid to the datasets used to benchmark different methods. 
It has been highlighted by Tello, et al. \cite{tello2024too} of inherent problems specific to HAR which is prone to overestimating the recognition performance.
Since human motion is highly repetitive in a short time but largely diverse across longer periods and individuals, the common practice of shuffling the datasets (which is sliced in small windows) and partitioning the training/testing split inherently introduces overfitting and does not represent performance when applied to new data, especially from unknown individuals.
Also holding out only selected individual or recording sessions as testing does not represent the entire dataset either. 
Instead, the proper practice is leaving out data from individuals or sessions in combination with cross-validation.
Since failing to adhere to this stringent data partition practice can lead to significant performance overestimation, we refrain from reiterating classification metrics reported by the literature and focus on the methodology.

\section{Methodology}


Classical benchmark procedures commonly aim to quantify the classification performance for a selected model architecture in order to promote superiority.
In this scope, the dataset is a means to an end to provide comparable results through shared training and validation data origins.
However, Beyond Confusion aims to investigate state-of-the-art HAR datasets independently of favorable benchmarking setups and machine learning model architectures.
Our sensor-data-oriented approach targets to analyze each dataset through the fusion of multi-model architecture training results to stress out the ambiguities of incorrectly classified sensor data. 
More specifically, we assume that the origin of the wrong classification has to lay in the dataset itself if the whole set of applied machine learning models cannot classify it properly.

\subsection{Datasets}
We selected six popular datasets from \cref{tab:related}: PAMAP2, Opportunity, MM-Fit, MHealth, MotionSense and WISDM to apply our methodology.
They present a large range of characteristics from the number of participants, type of activities, number of sensors, positions of sensors and their recording time.
Apart from the details in \cref{tab:related}, there are several notable nuances among the chosen datasets (extended details for each dataset in \cref{model_details}): 

\begin{itemize}
    \item The PAMAP2 dataset has effectively eight participants (the ninth was excluded by the original dataset) with three IMUs, positioned on the chest, dominant wrist, and dominant ankle. 
    Each participant provided one session of recording.
    Notably, one of the eight participants was left-handed and thus had a different sensor placement from the other right-handed participants. 
    It was performed in scripted sequences, and the annotation takes the form of the start and end time of each activity according to the experiment protocol which lasts a large chunk.

    \item The Opportunity dataset includes four participants each providing six recording sessions. In five sessions they performed unscripted daily activities freely and in one session a scripted 'drill run'. It was annotated on a fine granularity with different levels of contexts, with one label track for five locomotion classes, one track for five high-level activities (e.g. making breakfast), and one track for 18 atomic gestures (e.g. open drawer 2). We selected the locomotion and gesture tracks as they are taken as the usual benchmarks. 
    \item The Opportunity and MotionSense datasets contain multiple participants with multiple recordings, while PAMAP2, MHealth, WISDM have only one session per participant. MM-Fit has an unspecified session-user combination. Datasets with multiple sessions per user allow both leave-sessions-out and leave-persons-out splits.

\end{itemize}

All datasets have been pre-processed through the same procedure based on common practices \cite{deepconvlstm, suh2023tasked} to enhance comparability across datasets in order to generate a common ground based on their differences in original frequencies to achieve consistent sliding windows of 200 data points of the same length and 100 data points in stride for each.
For all datasets, the input data was normalized to zero mean and unity variance where mean and standard deviation were computed from training split.


\subsection{Models and Experiment Settings}
As a basis for our work, we aim to investigate the sections of false classifications of each dataset.
Therefore, we trained a set of selected deep learning model architectures that were once or currently considered the go-to ‘milestone' choices in HAR on those datasets to gather the list of false classifications, covering the landscape from a simple CNN architecture towards complex models based on transformer networks.

\begin{itemize}
    \item CNN: Many works have proposed convolutional blocks as feature encoders with fully connected layers as the classifier \cite{tang2020exploring:cnn,stromback2020mmfit,malekzadeh2018motionsense,yang2015deepcnn}. The convolution kernels and channels help to extract information across time steps and sensor channels.

    \item GRU: Gated Recurrent Units (GRUs) are recurrent neural networks that are specialized in encoding temporal information, thus it was adapted in \cite{haresamudram2022assessing}. 

    \item LSTM: Long Short-Term Memory (LSTM) models are essentially improved versions of GRUs with more parameters and gates to capture both long-term dependencies as well as subtle patterns \cite{lstm:convlstm,haresamudram2022assessing}.

    \item ConvLSTM: Combining convolution and LSTM layers was proposed in \cite{deepconvlstm,lstm:convlstm} specifically for HAR problems. 

    \item TinyHAR: Proposed in \cite{zhou2022tinyhar}, TinyHAR is a combination of CNN, transformer and LSTM layers, leveraging their distinct advantages and was shown to be superior to other neural networks currently in HAR benchmarks under proper leave-out cross-validation.
 
    \item CPC: Contrastive Predictive Coding (CPC) \cite{cpc:har,oord2018representation:cpc} is a model enhancement method through self-supervised pre-training of the feature encoder through a task of recurrently predicting the future latent features. CPC was selected among many self-supervised techniques both for brevity and its only requirement of continuous sensor data making it easily applicable to all of the datasets. The encoder was unlocked in the downstream classification task on the dataset, as it was shown to be more effective in \cite{fortes2023don}.  
    
\end{itemize}

Each model was trained with variations in learning rate and batch size to maximize the architecture's potential.
We executed the training with nine total combinations of common learning rate (0.1,0.01,0.001) and batch size (64,256,1024) selections.
Based on best practices, we added a learning rate decay with Gamma=0.5 and a step size of 20 epochs to support the model convergence.

Oppose to traditional benchmarking, with predefined dataset splits for training and validation, we applied Group-K-Fold as a cross-validation metric for the complete dataset.
Depending on the dataset, this strategy involved training separate models for each person or session independently by leaving out the selected one for validation and utilizing the rest for training.
For datasets with an increased number of sessions or persons, we decided to limit the number of folds to a maximum of 10 groups by merging them into larger groups.
However, the affiliation of wrongly classified windows can still be traced back by logging the ID of each.

Apart from the different model training setups, each model was trained with Adam optimizer and cross-entropy loss.
To terminate the training process without occupying too many resources for underperforming training runs, we applied an early stopping metric with a patience of 40 epochs.

The full training setup was executed four times with random weight initialization on two different multi-GPU setups running on NVIDIA H100 80GB and Nvidia RTX A6000 which each Dataset/Model combination shifted to one GPU.
Both systems were based on Ubuntu 20.04 with PyTorch 1.13.0 and the CUDA Toolkit 11.8.0.
For each training, we tracked the setup, the progress, and the final result in files for later investigation.
As discussed in the following sections, the results of the four total training runs were then merged and analyzed towards their deviations to obtain meaningful and repeatable results.


\subsection{Identifying Common False Classifications} 

Throughout the multiple dimensions of experiment setups with datasets, model architectures, hyperparameter selections, and the grouped leave-out cross-validation, we tracked the required data and information of each configuration next to the overall model training performance, especially the probabilities and data indices of the leave-out test set to reconstruct the training and evaluation process.
We extracted the best-performing hyperparameter setup from each dataset/model combination through a selection of the best final model accuracy as an initial step for all four full experiment runs.
As shown in \cref{tab:model_results_acc}, we calculated the mean accuracy across the four training runs and added the deviations.
The highlighted results represent the best performance for each model.
Additionally, the findings are reflected in \cref{tab:model_results_f1}, showing the mean weighted F1-Score with the according deviations.
We selected the weighted F1-score to properly account imbalance of datasets since the nature of HAR data from real study participants commonly involves increased distributions towards the null activity class.
Due to the proportional small deviations across our experiments, we can conclude the results as meaningful and repetitive.
As highlighted, the TinyHAR and CPC models outperform the less complex model architectures across all datasets.
However, apart from the PAMAP2 and the two Opportunity results, model architectures like CNN were able to keep up with the best benchmarking results with sometimes less than a few percent difference.

\begin{table}[!t]
\footnotesize
\centering
\caption{Accuracy (\%) for each Model/Dataset Combination (mean ± std.)}
\begin{tabular}{ccccccc}
\toprule
& \textbf{CNN} & \textbf{CONVLSTM} & \textbf{GRU} & \textbf{LSTM} & \textbf{TinyHAR} &\textbf{CPC} \\
\midrule
\textbf{PAMAP2} & 65.83 ± 4.07 & 70.84 ± 4.12 & 65.94 ± 3.08 & 65.95 ± 3.96 & \textbf{74.86 ± 3.29} & 63.58 ± 3.08 \\
\textbf{Oppo-Loco} & 70.67 ± 3.12 & 70.54 ± 2.01 & 74.72 ± 2.63 & 70.39 ± 2.11 & \textbf{82.36 ± 2.47} & 71.44 ± 2.00 \\
\textbf{Oppo-Gest} & 75.99 ± 3.32 & 72.86 ± 3.68 & 77.76 ± 2.47 & 75.58 ± 2.79 & \textbf{79.55 ± 2.33} & 75.67 ± 2.62 \\
\textbf{MM-FIT} & 95.95 ± 2.87 & 97.17 ± 2.03 & 96.70 ± 2.76 & 95.89 ± 2.62 & \textbf{97.97 ± 1.58} & 96.73 ± 1.76 \\
\textbf{MHEALTH} & 80.91 ± 2.43 & 79.75 ± 2.39 & 81.27 ± 2.51 & 78.53 ± 2.26 & \textbf{83.41 ± 2.50} & 80.94 ± 2.35 \\
\textbf{MotionSense} & 98.27 ± 2.95 & 97.32 ± 2.09 & 97.89 ± 2.03 & 97.94 ± 1.84 & 98.08 ± 1.73 & \textbf{98.92 ± 1.46} \\
\textbf{WISDM} & 88.90 ± 2.67 & 89.31 ± 2.81 & 90.62 ± 2.44 & 88.19 ± 2.36 & 91.78 ± 2.55 & \textbf{92.94 ± 2.59} \\
\bottomrule
\end{tabular}
\label{tab:model_results_acc}
\end{table}

\begin{table}[!t]
\footnotesize
\centering
\caption{Weighted F1 Scores (\%) for each Model/Dataset Combination (mean ± std.)}
\begin{tabular}{ccccccc}
\toprule
& \textbf{CNN} & \textbf{CONVLSTM} & \textbf{GRU} & \textbf{LSTM} & \textbf{TinyHAR} &\textbf{CPC} \\
\midrule
\textbf{PAMAP2} & 64.73 ± 5.54 & 69.64 ± 4.11 & 64.12 ± 4.04 & 64.34 ± 4.62 & \textbf{73.83 ± 3.37} & 68.82 ± 4.13 \\
\textbf{Oppo-Loco} & 70.57 ± 3.53 & 71.06 ± 3.11 & 74.79 ± 2.66 & 70.52 ± 2.98 & \textbf{82.21 ± 2.53} & 71.30 ± 2.46 \\
\textbf{Oppo-Gest} & 71.97 ± 3.11 & 61.24 ± 4.41 & 74.40  ± 2.72 & 70.49 ± 2.36 & \textbf{76.69 ± 2.11} & 72.25  ± 2.28 \\
\textbf{MM-FIT} & 95.91 ± 2.91 & 97.16 ± 2.18 & 96.67 ± 2.84 & 95.89 ± 2.63 & \textbf{97.95 ± 1.60} & 96.67 ± 1.45 \\
\textbf{MHEALTH} & 80.30 ± 2.40 & 79.58 ± 2.84 & 79.61 ± 2.98 & 76.83 ± 3.07 & \textbf{82.66 ± 2.41} & 79.90 ± 2.77 \\
\textbf{MotionSense} & 98.30 ± 3.79 & 97.37 ± 2.15 & 97.92 ± 2.06 & 97.97 ± 2.80 & 98.13 ± 1.73 & \textbf{98.89 ± 1.55} \\
\textbf{WISDM} & 89.04 ± 2.63 & 89.52 ± 3.01 & 90.52 ± 2.91 & 88.22 ± 2.76 & 91.89 ± 1.53 & \textbf{93.01 ± 1.67} \\
\bottomrule
\end{tabular}
\label{tab:model_results_f1}
\end{table}

Opposed to reaching the maximum of 100\% accuracy, we can derive slight trends for a critical limit for each dataset, from where no significant improvement occurred.
Therefore, we conducted an overlap calculation for each dataset to extract the false classified windows from our evaluations.
We assume that for each examined dataset, there is a threshold value that indicates the percentile distribution between clean and issue-free data versus data that potentially contains disturbances in sensor signal quality or possible inaccuracies within the ground truth labels.
The Intersect of False Classifications (IFC) therefore represents the parts of the dataset, where none of the models could classify the window correctly.
If one is able to classify the sensor signal properly, it is an indicator that such a section of the dataset contains sufficient quality to be classified.

\begin{figure}[!b]
\footnotesize
    \centering
    \includegraphics[width=\textwidth]{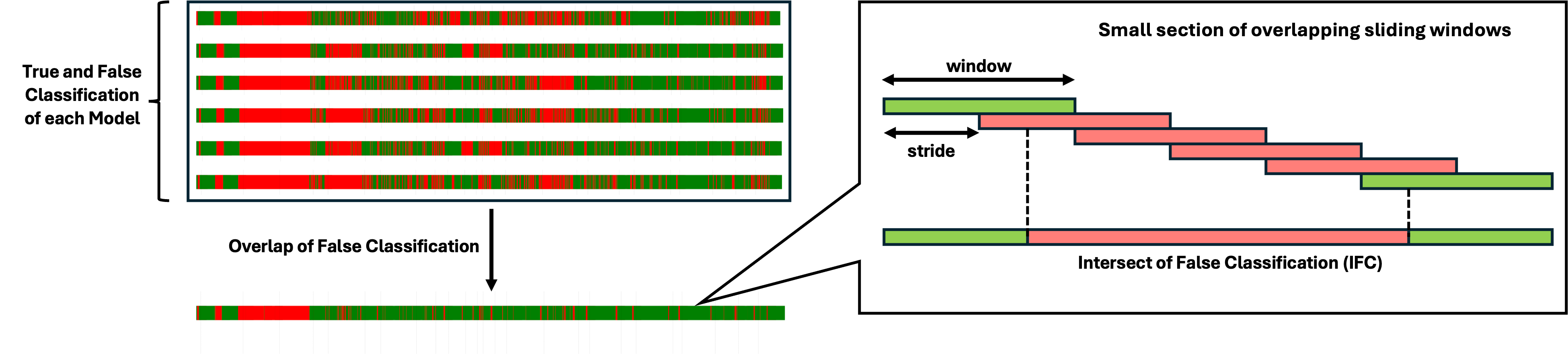}
    \caption{Extracting the Overlap of False Classification across the models, then merging the overlapping slinging windows into the 1D sequence for the Intersect of False Classification (IFC).}
    \Description{}
    \label{fig:IFC}
\end{figure}

We establish the IFC as a metric to quantify the amount of misclassifications as shown in \Cref{fig:IFC} by extracting the false classifications across the models and merging the sliding-window-based
Across each model per dataset, we extracted the percentage of windows that were only classified correctly by one of the respective models as shown in \cref{tab:false_classes}.
Although one would assume that less-performing models contain equal subsets of false classified windows from the most performing models, however, each of the architectures could still contribute a small portion to the correct classified windows.
This however manifests our idea of calculating the IFC across the models to establish the dataset analysis independently from the model architectures.
Together with the common ground, quantifying the overlap of correct classification from multiple models, we calculate the IFC through the following \cref{eq:IFC}

\begin{equation}
  \text{IFC} = 100\% - \text{Common Ground} - \sum^{\text{models}}(\text{Single Contributions})
  \label{eq:IFC}
\end{equation}

It is noteworthy to distinguish the IFC from the maximum achievable performance.
Moreover, this metric represents a measure of the cleanness and suitability of the data, since our approach targets the dataset instead of the model architecture.
For overly large and complex models, predominantly developed for benchmarking purposes, combined with advanced or specialized preprocessing approaches, the achieved model performance may excel in the correct classification beyond our findings since they can adapt and therefore diminish the effects of data ambiguities.
However, they may not suit our approach to quantify the dataset quality and stress out the ambiguities.

For each dataset, we investigate the cause of the false classified data included in the IFC to highlight potential issues when generating a HAR dataset.
This is mainly the reason, why each dataset was fully exposed in our experiments by training it on all available persons, sessions, and the full set of IMU data streams.

\begin{table}[!t]
\footnotesize
  \centering
  \caption{Isolated contribution of each model next to the shared common ground across all models to calculate the Intersect of False Classifications (IFC).}
  \begin{tabular}{ccccccc|c|c}
    \toprule
     & \textbf{CNN} (\%)& \textbf{CONVLSTM} (\%)& \textbf{GRU} (\%)& \textbf{LSTM} (\%)& \textbf{tinyHAR} (\%)&\textbf{CPC} (\%)& \thead{{\textbf{Common}} \\ {\textbf{Ground}} {(\%)}} & \textbf{IFC} (\%)\\ 
    \midrule
     \textbf{PAMAP2} & 0.72 & 1.42 & 0.76 & 0.96 & 3.04 & 0.59 & 80.77 & 11.74\\  
    \textbf{Oppo-Loco} & 0.60 & 0.78 & 0.78 & 0.64 & 1.79 & 1.25 & 88.85 & 5.31\\ 
    \textbf{Oppo-Gest} &0.29 &0.20 & 1.20 & 0.30 & 2.01 & 0.91& 84.23 & 12.06\\ 
    \textbf{MM-FIT} & 0.06 & 0.13 & 0.10 &  0.05 & 0.19 & 0.06 & 98.97 & 0.44\\  
    \textbf{MHEALTH} & 0.85 & 1.52 & 1.05 & 0.82 & 1.19 & 0.68 & 89.31 & 4.58\\  
    \textbf{MotionS.} & 0.03 & 0.08 & 0.01 & 0.01 & 0.16 & 0.10  & 99.15 & 0.46\\  
    \textbf{WISDM} & 0.27 & 0.15 & 0.25 & 0.17 & 0.41 & 0.31 & 96.83 & 1.61\\  
    \bottomrule
  \end{tabular}
  \label{tab:false_classes}
  
\end{table}

\section{Contextual Analysis}
\label{sec:statistics}

Compared to the common evaluation of deep learning model training results, traditionally presenting the metrics of the full benchmarking dataset setup, we target to shift the model performance results towards an in-depth dataset analysis to carve out the cause for the false classifications.
Therefore, this section presents the process of thoroughly investigating the experiment results, especially the Intersect of False Classifications.
With such an approach, we can narrow the source of false classification down to the raw sensor signal and ground truth labels.

\subsection{False Classes Identification}

To meaningfully interpret the IFC, we decided to elaborate on the falsely classified windows in more detail by extracting the correct and confused predictions next to the amount of confusion from our evaluations.
Since false classified windows may be assigned different classes by each model, we did not utilize the classes but rather the probability distribution of each class.
Therefore, the results presented in this section rest upon the mean calculation of false classified probability distributions across the trained models, followed by selecting the confused class by maximum portability selection.
Again, this approach leads to a model-independent analysis of each dataset by fusing the evaluation results.
For each dataset, we visualize the confused classes in the form of chord diagrams.
Such circular visualization obtains the set of classes, distributed in a circular arrangement, whereas arrows connect the classes between each other.
An arrow within the diagram always points from the true class to the confused class, highlighted with the true class color.
The thickness of each arrow additionally represents the ratio of confusion occurrence in proportion to the other arrows.
Conclusions towards the false classes distribution can therefore easily be drawn since increased occurrence of confusion stands out in the diagram.
Additionally, to interpret the findings in a more quantifiable way, we added a table for each dataset, showing first the distribution of ground truth classes in the dataset, and secondly the relative and absolute confusion.
In this context, relative confusion states the percentage of confusion within the corresponding class, whereas absolute represents the confusion across the whole dataset.

\begin{figure}[!t]
    \begin{subfigure}{0.45\textwidth}
        \centering
        \includegraphics[width=\textwidth]{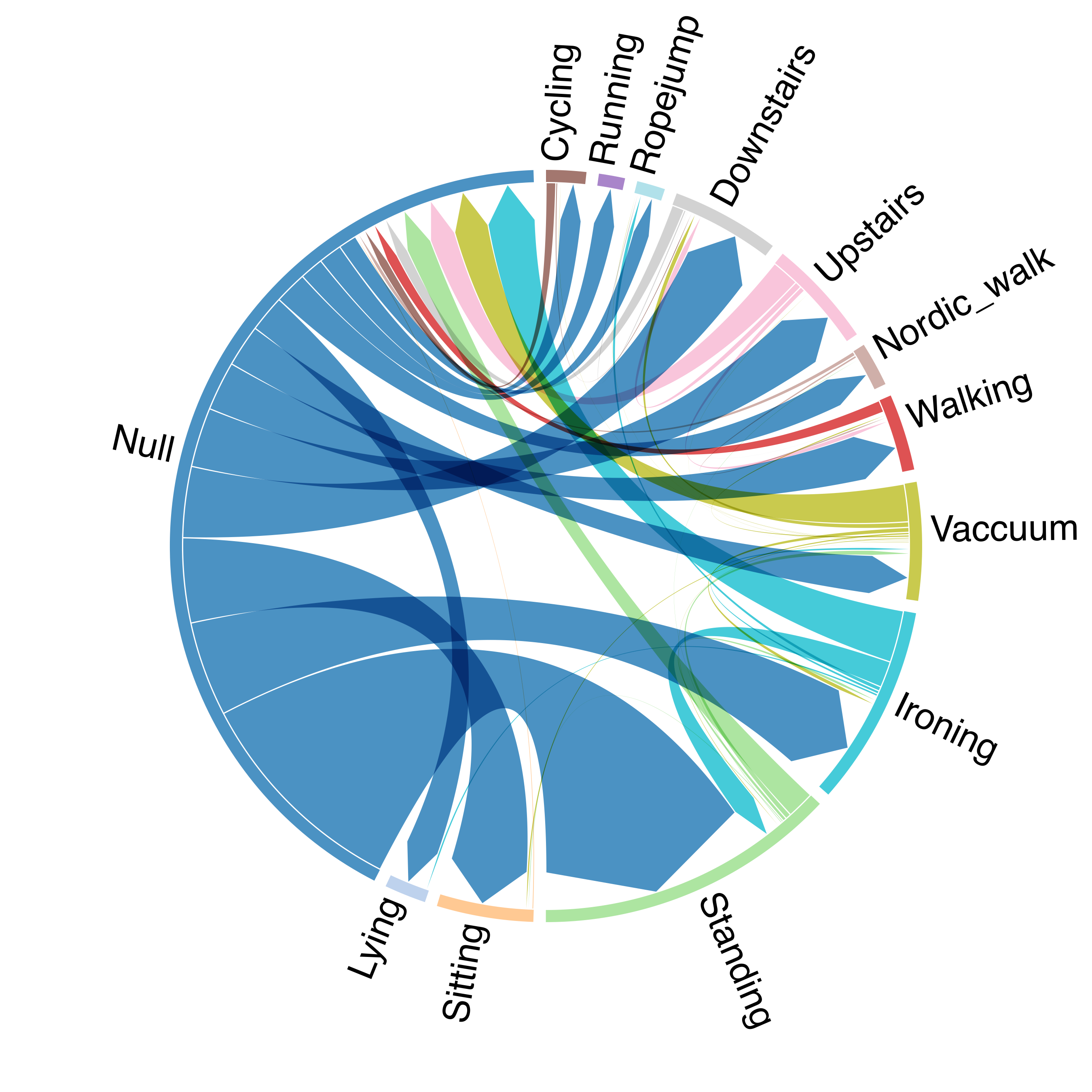}
    \end{subfigure}
    \hfill
    \begin{subfigure}{0.45\textwidth}
        \centering
        \includegraphics[width=\textwidth]{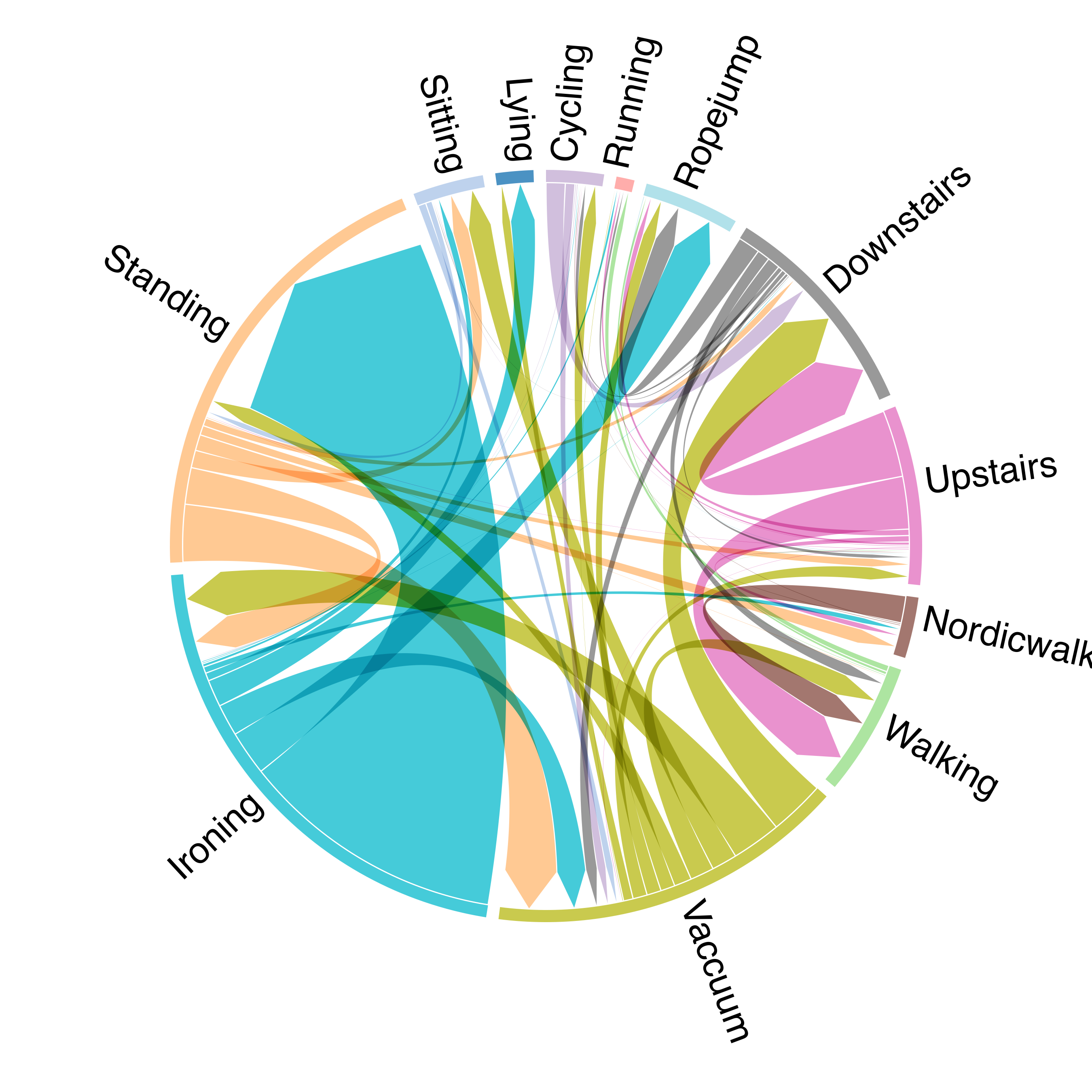}
        
    \end{subfigure}
    \caption{Chord diagrams of PAMAP2, with \textit{null} class added on the left and \textit{null} class removed on the right for clarity.}
    \Description{}
    \label{fig:pamap2}
\end{figure}

\begin{table}[!t]
\footnotesize
\centering
\caption{Class distribution and confusion in PAMAP2.}
\begin{tabular}{ccccc}
\toprule
\textbf{ID} & \textbf{True Class} & \textbf{Dist. (\%)} & \textbf{Conf. Rel. (\%)} & \textbf{Conf. Abs. (\%)} \\ 
\midrule
0 & null & 32.37 & 21.73 & 8.41\\
1 & lying & 6.72 & -  & -\\ 
2 & sitting & 6.47 & 0.67 & 0.03 \\ 
3 & standing & 6.63 & 10.82 & 0.47 \\ 
4 & walking & 8.34 & 1.88 & 0.15\\ 
5 & running & 3.43 & - & - \\ 
6 & cycling & 5.75 & 2.55 & 0.14 \\ 
7 & nordic\_walking & 6.57 & 1.31 & 0.08 \\ 
8 & up\_stairs & 4.09 & 12.03 & 0.46\\ 
9 & down\_stairs & 3.66 & 7.13 & 0.21 \\ 
10 & vaccuum\_cleaning & 6.13 & 10.75 & 0.67 \\ 
11 & ironing & 8.34 & 12.28 & 1.01\\ 
12 & rope\_jumping & 1.50 & 0.15 & 0.002\\ 
\bottomrule
\end{tabular}
\label{tab:pamap2_dist}
\end{table}

Starting with the PAMAP2 dataset, \cref{fig:pamap2} shows an increased confusion of the \textit{null} class, with almost half of the circular arrangement occupying.
Supported by \cref{tab:pamap2_dist}, the \textit{null} class obtains over 21\% relative confusion.
Together with the overly large distribution of 32\%, the \textit{null} class contributes over 8\% to the absolute confusion, whereas the other classes are consistently below 1\%.
One explanation for this could be the labeling design used, which could possibly contain sensor signals from concurrent activities.

We additionally added a chord diagram in \cref{fig:pamap2} excluding the \textit{null} class to highlight the activities.
The confusion between ironing with standing is particularly striking in this visualization.
Depending on the experiment design, ironing commonly includes standing when utilizing an ironing board.
Presumably, the arm movement for ironing does not suffice to distinguish the activities properly.
For single output classification problems, the labels are supposed to be mutually exclusive from each other.
Thus the mixture of higher context level activities such as ironing and vacuuming which can contain activities like standing or walking could introduce ambiguity.


\begin{figure*}[!t]
\centering
\begin{adjustbox}{valign=c}
  \begin{subfigure}[b]{0.45\textwidth}
  \includegraphics[width=\textwidth]{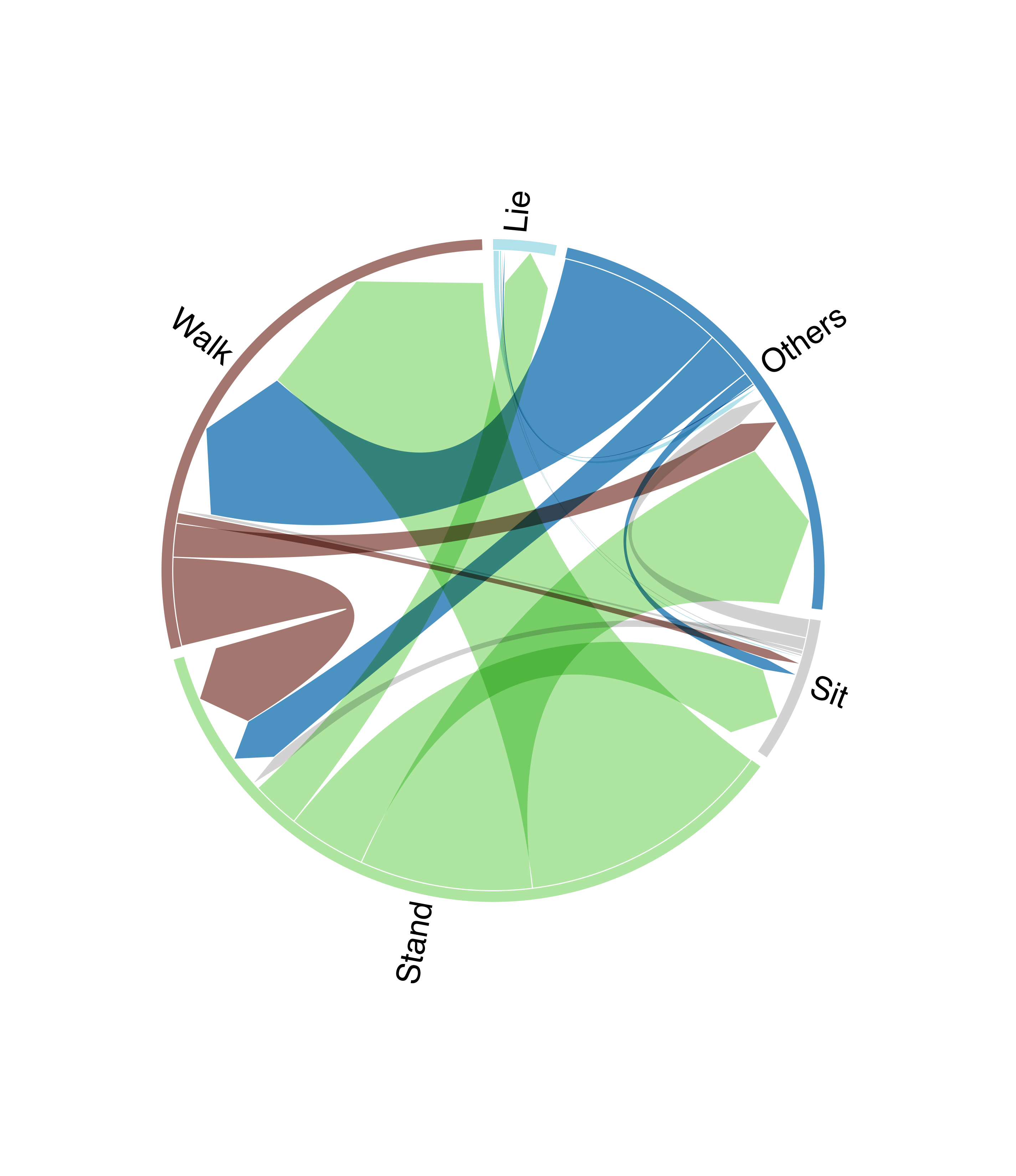}
  \end{subfigure}
\end{adjustbox}
\begin{adjustbox}{valign=c}
  \begin{subtable}[b]{0.5\textwidth}
  \centering
  \footnotesize
  \begin{tabular}{ccccc}
\toprule
\textbf{ID} &\textbf{True Class} & \textbf{Dist. (\%)} & \textbf{Conf. Rel. (\%)} & \textbf{Conf. Abs. (\%)} \\ 
\midrule
0 & Others & 17.96 & 7.03 & 1.26\\
1 & Stand & 41.58 & 6.97  & 3.06\\ 
2 & Walk & 21.29 & 3.71  & 0.73 \\ 
3 & Sit & 16.14 & 1.32  & 0.21 \\ 
4 & Lie & 3.03 & 1.66  & 0.05\\ 
\bottomrule
\end{tabular}
  \end{subtable}
\end{adjustbox}
\caption{Chord diagrams and table of class distribution and confusion in Opportunity with locomotion labels.}
\Description{}
\label{fig:Oppo_loco}
\end{figure*}

\begin{figure*}[!t]
\centering
\begin{adjustbox}{valign=c}
  \begin{subfigure}[b]{0.45\textwidth}
  \includegraphics[width=\textwidth]{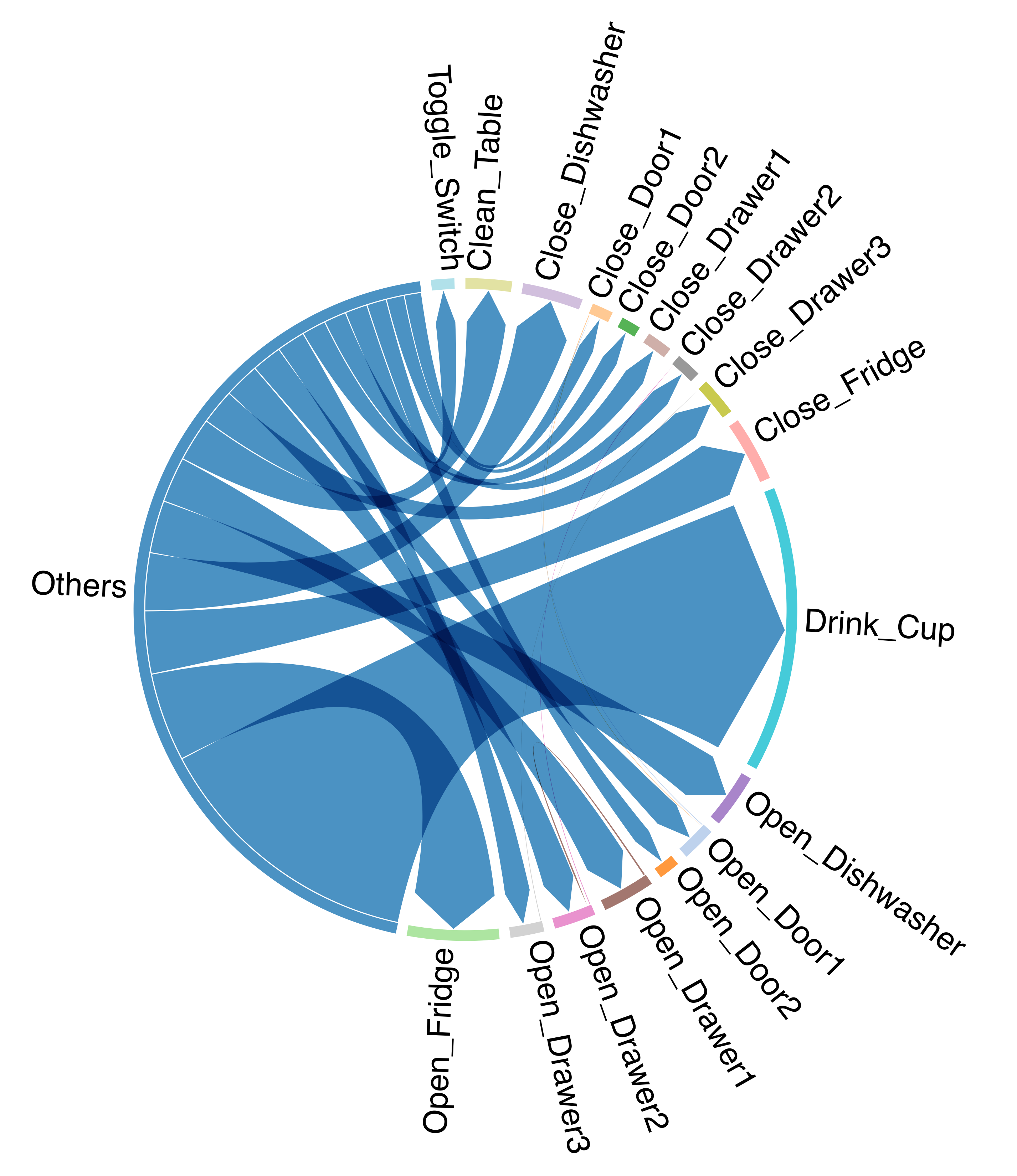}
  \end{subfigure}
\end{adjustbox}
\begin{adjustbox}{valign=c}
  \begin{subtable}[b]{0.54\textwidth}
  \centering
  \footnotesize
  \begin{tabular}{ccccc}
\toprule
\textbf{ID} &\textbf{True Class} & \textbf{Dist. (\%)} & \textbf{Conf. Rel. (\%)} & \textbf{Conf. Abs. (\%)} \\ 
\midrule
0 & Others & 72.73 & 13.99 & 11.83 \\
1 & Open\_Door1 & 1.64 & 1.87 & 0.02 \\
2 & Open\_Door2 & 1.60 & - & - \\
3 & Close\_Door1 & 1.69 & 1.59 & 0.02 \\
4 & Close\_Door2 & 1.72 & 0.77 & 0.01 \\
5 & Open\_Fridge & 1.744 & - & - \\
6 & Close\_Fridge & 0.94 & - & - \\
7 & Open\_Dishwasher & 1.29 & - & - \\
8 & Close\_Dishwasher & 0.97 & - & - \\
9 & Open\_Drawer1 & 0.85 & 20.83 & 0.58 \\
10 & Close\_Drawer1 & 0.34 & - & - \\
11 & Open\_Drawer2 & 1.04 & 6.12 & 0.035 \\
12 & Close\_Drawer2 & 0.34 & - & - \\
13 & Open\_Drawer3 & 1.22 & 5.33 & 0.05 \\
14 & Close\_Drawer3 & 0.74 & - & - \\
15 & Clean\_Table & 1.81 & 0.94 & 0.01 \\
16 & Drink\_Cup & 9.37 & - & - \\
17 & Toggle\_Switch & 0.30 & - & - \\
\bottomrule
\end{tabular}
  \end{subtable}
\end{adjustbox}
\caption{Chord diagrams and table of class distribution and confusion in Opportunity with gesture labels.}
\Description{}
\label{fig:Oppo_gest}
\end{figure*}

Since the Opportunity dataset is labeled with two tracks of activity classes for the same data, we conducted experiments for the locomotion and the gesture labels.
The findings are presented in \cref{fig:Oppo_loco} and \cref{fig:Oppo_gest}.
As a specialty for this dataset, either the locomotion or the gesture labels contain identical sensor streams, whereas the \textit{others} class increasingly contains the section from the other labeling set.
For instance, when there is locomotion like walking in the data, there is less gesture interaction possible to open or close any items.

For the case of locomotion in \cref{fig:Oppo_loco}, \textit{standing} obtains an increased confusion with \textit{walking}.
Even though these two activities should be distinguishable properly, \textit{standing} may not correlate with \textit{standing still}.
Due to the two labeling streams, \textit{standing} could include parts of the applied gesture activities, leading to confusion with walking due to the motion.

The majority of the Opportunity dataset contains locomotion data, whereas the \textit{others} class for gestures obtains a large distribution of over 72\% as shown in \cref{fig:Oppo_gest}.
Especially when examining the absolute confusion, the \textit{others} class rules the overall classification accuracy since multiple different activities from the locomotion are merged into one class.
Oppose that, the activity classes are distinguished properly with some of them being never confused.

\begin{figure*}[!t]
\centering
\begin{adjustbox}{valign=c}
  \begin{subfigure}[b]{0.45\textwidth}
  \includegraphics[width=0.9\textwidth]{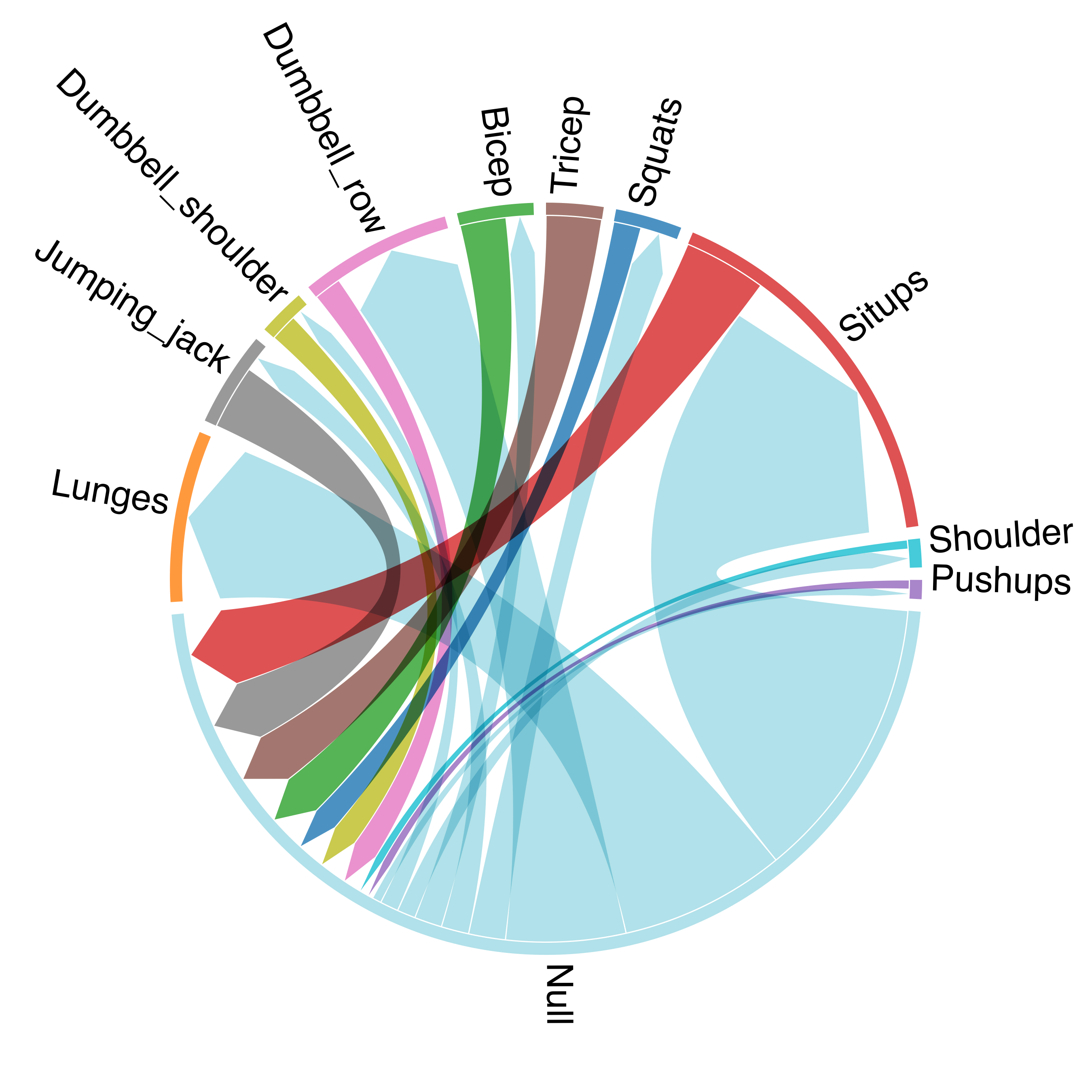}
  \end{subfigure}
\end{adjustbox}
\begin{adjustbox}{valign=c}
  \begin{subtable}[b]{0.5\textwidth}
  \centering
  \footnotesize
  \begin{tabular}{ccccc}
\toprule
\textbf{ID} &\textbf{True Class} & \textbf{Dist. (\%)} & \textbf{Conf. Rel. (\%)} & \textbf{Conf. Abs. (\%)} \\ 
\midrule
0 &Squats & 2.68 & 0.43 & 0.011 \\
1 &Lunges & 3.60 & - & - \\
2 &Bicep& 2.39 & 0.80 & 0.019 \\
3 &Situps & 3.31 & 1.06 & 0.034 \\
4 &Pushups & 2.19 & 0.17 & 0.004 \\
5 &Tricep & 2.64 & 0.86 & 0.023 \\
6 & Dumbbell Rows & 2.32 & 0.50 & 0.011 \\
7 &Jumping Jacks & 1.24 & 2.13 & 0.027 \\
8 &Dumbbell Shoul. & 2.78 & 0.41 & 0.011 \\
9 &Shoulder & 2.68 & 0.14 & 0.004 \\
10 &null & 74.18 & 0.39 & 0.29 \\
\bottomrule
\end{tabular}
  
  \end{subtable}
\end{adjustbox}
\caption{Chord diagrams and table of class distribution and confusion in MM-FIT.}
\Description{}
\label{fig:mmfit}
\end{figure*}



For the MM-FIT dataset, \cref{fig:mmfit} reflects a major imbalance towards the \textit{null} class with over 74\%, accounting for the majority of the confusion.
Even though the relative confusion is low for the \textit{null} class, the absolute confusion rules the prediction performance.
Again, since the \textit{null} class is dominantly confused with \textit{lunges} and \textit{sit ups} as shown in the chord diagram, it raises the question if those activities include similar sensor information as the windows labeled with \textit{null} class.
On the contrary, even though relative confusion for other activity classes like \textit{jumping jacks} is increased, it is negligible in the total confusion due to the dominance of the \textit{null} class.

\begin{figure*}[!t]
\centering
\begin{adjustbox}{valign=c}
  \begin{subfigure}[b]{0.45\textwidth}
  \includegraphics[width=0.9\textwidth]{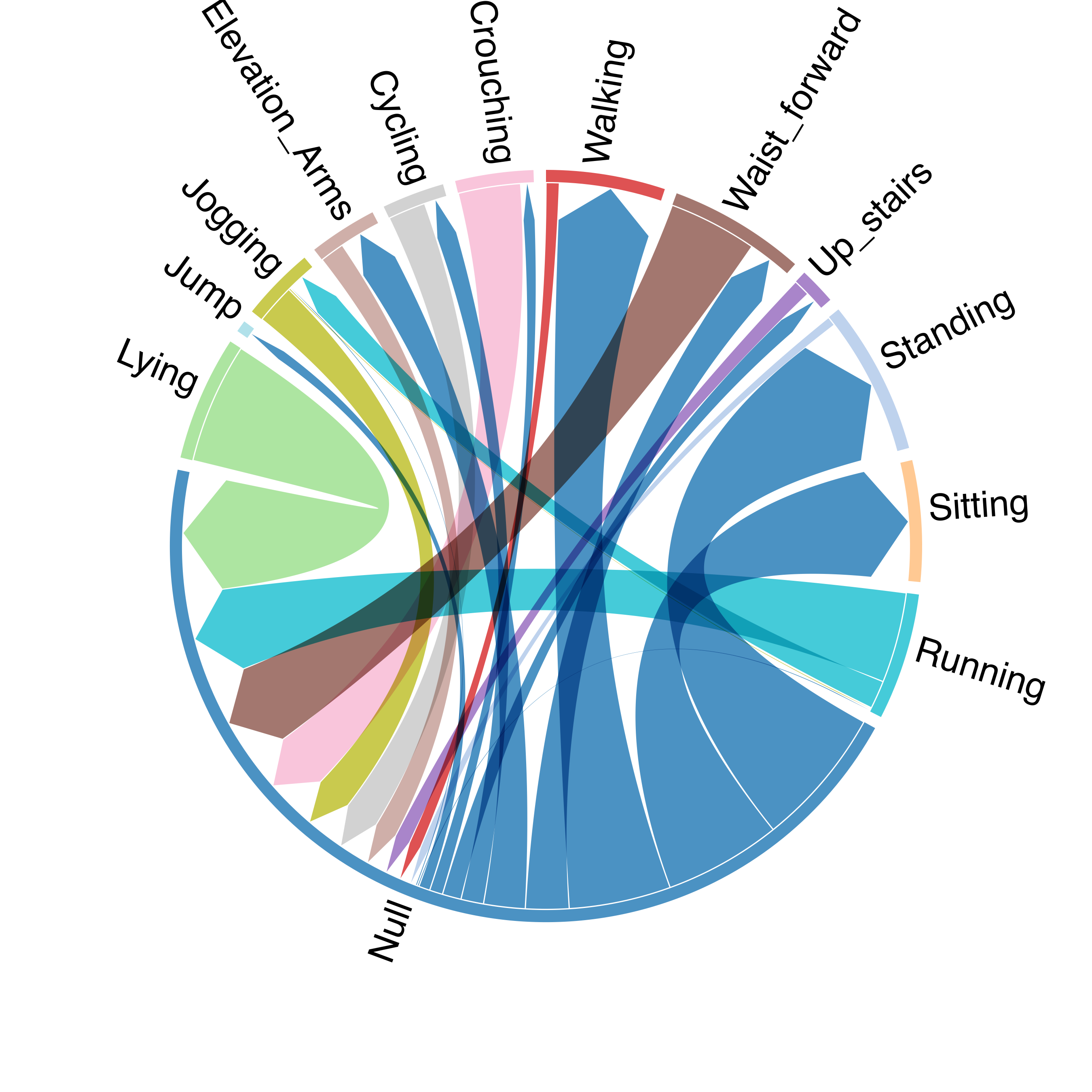}
  \end{subfigure}
\end{adjustbox}
\begin{adjustbox}{valign=c}
  \begin{subtable}[b]{0.5\textwidth}
  \centering
  \footnotesize
  \begin{tabular}{ccccc}
  \toprule
  \textbf{ID} &\textbf{True Class} & \textbf{Dist. (\%)} & \textbf{Conf. Rel. (\%)} & \textbf{Conf. Abs. (\%)} \\ 
  \midrule
  0 & null & 71.77\% & 3.119 & 2.240 \\
  1 & Standing still & 2.52\% & 2.110 & 0.041 \\
  2 & Sitting & 2.53\% & - & - \\
  3 & Lying & 2.54\% & 17.158 & 0.527 \\
  4 & Walking & 2.54\% & 2.672 & 0.058 \\
  5 & Upstairs & 2.50\% & 2.970 & 0.074 \\
  6 & Waist forward & 2.32\% & 15.359 & 0.387 \\
  7 & Elevation Arms & 2.44\% & 4.530 & 0.107 \\
  8 & Crouching & 2.41\% & 10.345 & 0.272 \\
  9 & Cycling & 2.52\% & 6.369 & 0.165 \\
  10 & Jogging & 2.51\% & 7.074 & 0.181 \\
  11 & Running & 2.52\% & 17.166 & 0.519 \\
  12 & Jump & 0.86\% & - & - \\
  \bottomrule
  \end{tabular}

  \end{subtable}
\end{adjustbox}
\caption{Chord diagrams and table of class distribution and confusion in MHealth.}
\Description{}
\label{fig:mhealth}
\end{figure*}

Checking the relative confusion of the MHealth dataset in \cref{fig:mhealth}, multiple classes obtain increased percentage values of up to 17\%, for instance, the classes \textit{lying down} and \textit{running}. 
However, the same experiment design like MM-FIT leads to an overly large occurrence of \textit{null} classes with over 70\%.
Even though the relative confusion within the null is only slightly above 3\%, the absolute classification performance for MM-FIT-based models mainly depends on the \textit{null} activity class again.
For the remaining activities, the absolute confusion for each of them is located in the low percentage range below 1\%.
Additionally, when inspecting the chord diagram, confusion mainly occurs between the activity classes and the \textit{null} class, whereas apart from minimal confusion, the activities can be distinguished properly.
Checking the blue arrow for distinguishing the pure binary detection if or if not an activity is currently conducted, there is a lack of clear separation.

\begin{figure*}[!t]
\centering
\begin{adjustbox}{valign=c}
  \begin{subfigure}[b]{0.45\textwidth}
  \includegraphics[width=0.9\textwidth]{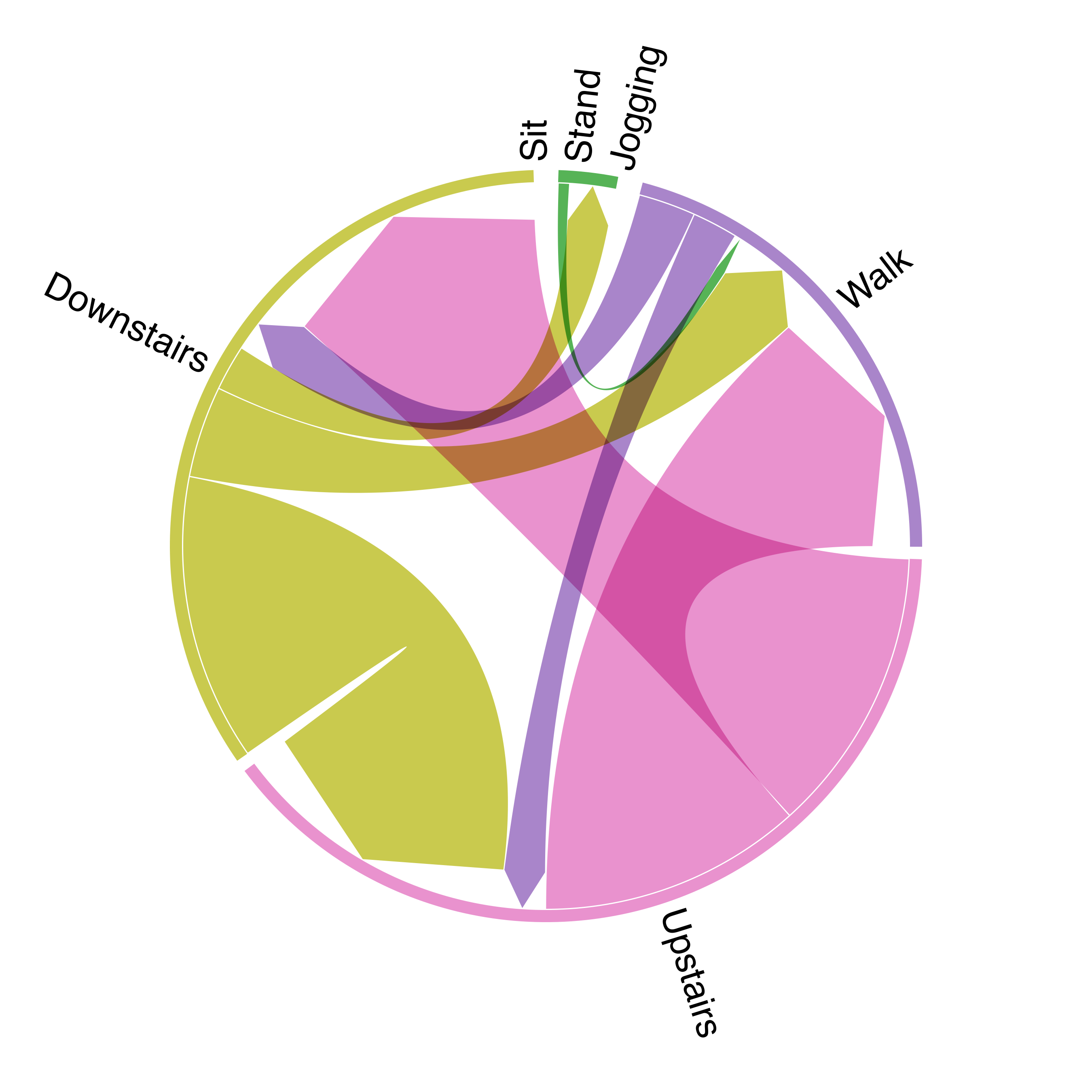}
  \end{subfigure}
\end{adjustbox}
\begin{adjustbox}{valign=c}
  \begin{subtable}[b]{0.5\textwidth}
  \footnotesize
\begin{tabular}{ccccc}
\toprule
\textbf{ID} &\textbf{True Class} & \textbf{Dist. (\%)} & \textbf{Conf. Rel. (\%)} & \textbf{Conf. Abs. (\%)} \\ 
\midrule
0 & Sit & 24.25 & - & - \\
1 & Stand & 22.42 & 0.021 & 0.005 \\
2 & Walk & 24.45 & 0.157 & 0.038 \\
3 & Upstairs & 10.77 & 2.109 & 0.229 \\
4 & Downstairs & 8.84 & 1.988 & 0.177 \\
5 & Jogging & 9.26 & - & - \\
\bottomrule
\end{tabular}

  \end{subtable}
\end{adjustbox}
\caption{Chord diagrams and table of class distribution and confusion in MotionSense.}
\Description{}
\label{fig:motion}
\end{figure*}

Due to a different experiment conducted with a clear separation of activities and fine-grained labeling of activities, the MotionSense dataset does not obtain a \textit{null} class.
As already determined in the previous section, MotionSense achieved great classification performance with over 98\% accuracy and weighted F1-Score.
Since there are not many false classified windows available for this dataset together and the set of activities obtains only six classes, the thickness of arrows in the chord diagram leads to the impression of major false classifications between classes.
With a focus on the table in \cref{fig:motion}, two of the six classes had no confusion across the whole experiment and the remaining classes obtained low absolute confusion, which is why the chord diagram appears convoluted.
The comparatively greatest confusions occur between \textit{upstairs}, \textit{downstairs}, and \textit{walking}, which is in the nature of things, as these activities all represent a type of uniform leg movement.

\begin{figure*}[!t]
\centering
\begin{adjustbox}{valign=c}
  \begin{subfigure}[b]{0.45\textwidth}
  \includegraphics[width=0.9\textwidth]{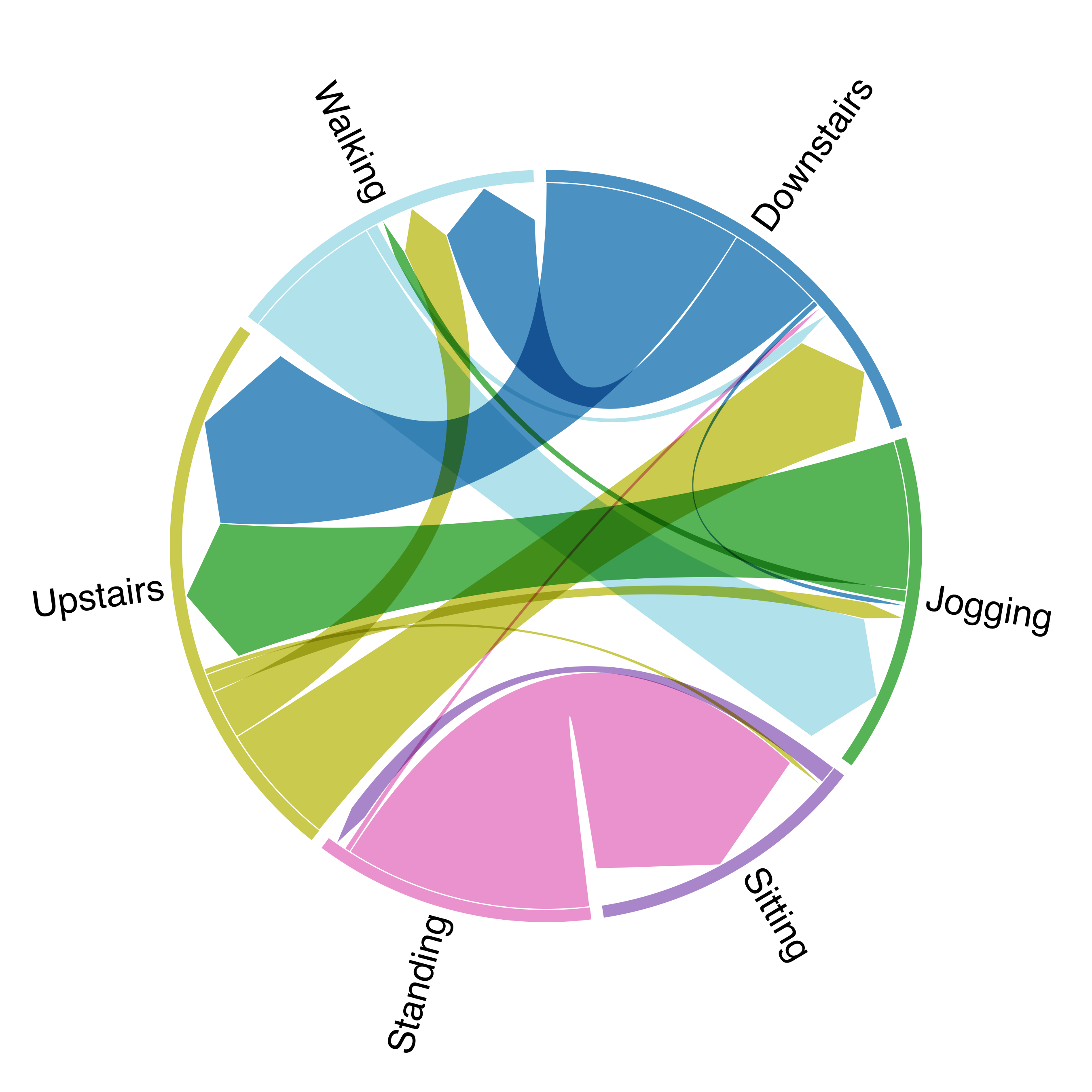}
  \end{subfigure}
\end{adjustbox}
\begin{adjustbox}{valign=c}
  \begin{subtable}[b]{0.5\textwidth}
  \footnotesize
  \begin{tabular}{ccccc}
\toprule
\textbf{ID} &\textbf{True Class} & \textbf{Dist. (\%)} & \textbf{Conf. Rel. (\%)} & \textbf{Conf. Abs. (\%)} \\ 
\midrule
0 & Downstairs & 9.15 & 4.776 & 0.448 \\
1 & Jogging & 31.31 & 0.760 & 0.238 \\
2 & Sitting & 5.43 & 0.541 & 0.027 \\
3 & Standing & 4.27 & 8.119 & 0.375 \\
4 & Upstairs & 11.20 & 2.667 & 0.293 \\
5 & Walking & 38.64 & 0.592 & 0.229 \\
\bottomrule
\end{tabular}
\

  \end{subtable}
  \end{adjustbox}
\caption{Chord diagrams and table of class distribution and confusion in WISDM.}
\Description{}
\label{fig:wisdm}
\end{figure*}

For the WISDM dataset, our experiments showed satisfying performance with accuracy and F1-Score around 90\%.
The distribution of classes and the relative confusion correlate across the activity classes, visualized in \cref{fig:wisdm}, showcases the effects of insufficient data quantity for underrepresented activities.
Especially for the classes \textit{downstairs} and \textit{standing}, the low representation in the dataset leads to an increased percentage of relative confusion.
Even though the total absolute confusion still maintains below 1\% for each class, an improved distribution of activity classes would lead to enhanced classification performance.
From \cref{fig:wisdm}, we can group the activities into stationary and mobile, with stationary containing \textit{standing} and \textit{sitting} and mobile the remaining activities.
Both of the groups are not confused due to the significant differences in the nature of activities, whereas within each group, confusion mainly occurs due to insufficient information from only the one smartphone accelerator, for instance, the confusion of \textit{walking} and \textit{jogging}.

In summary, we primarily recognize issues with class imbalance, especially dominance, and the associated confusion with targeted activity classes.
Beyond the difficulties from experiment conduction and the labeling granularity, primarily influencing the sensor data and ground truth quality, the set of designated activities can affect confusion when movements or poses can be mixed due to their common nature in physiological execution.

\subsection{Duration of False Classifications}
\label{sec:histograms}
\begin{figure}[!t]
    \centering
    \begin{subfigure}{0.24\textwidth}
        \centering
        \includegraphics[width=\textwidth]{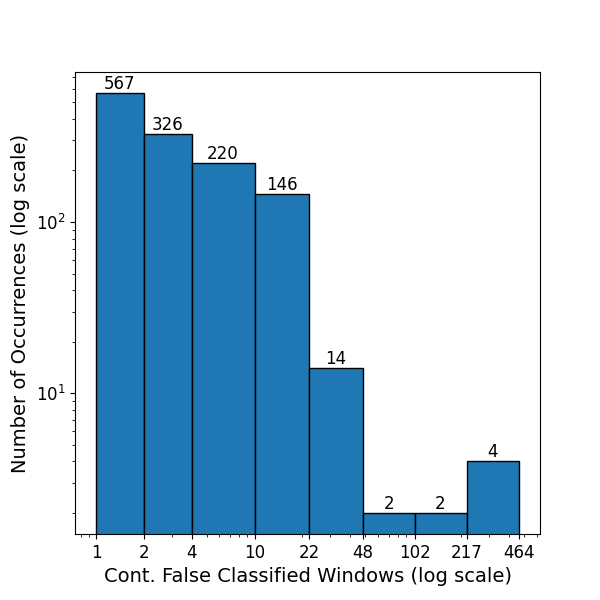}
        \caption{PAMAP2}
        \label{fig:sub1}
    \end{subfigure}
    \hfill
    \begin{subfigure}{0.24\textwidth}
        \centering
        \includegraphics[width=\textwidth]{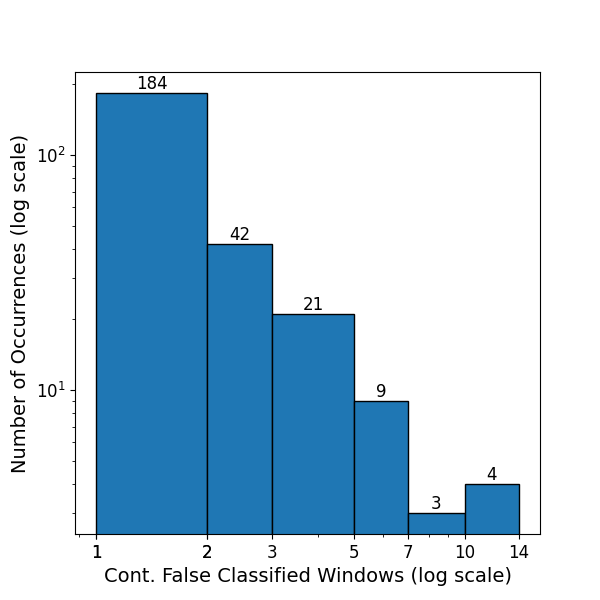}
        \caption{Opportunity Locomotion}
        \label{fig:sub2}
    \end{subfigure}
    \hfill
    \begin{subfigure}{0.24\textwidth}
        \centering
        \includegraphics[width=\textwidth]{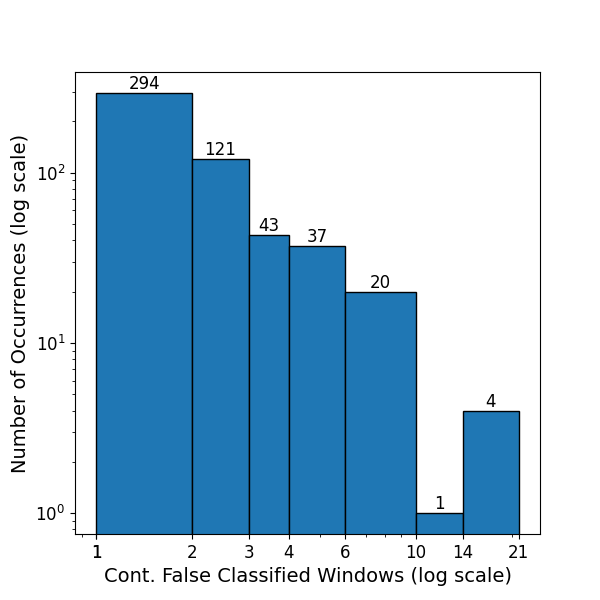}
        \caption{Opportunity Gesture}
        \label{fig:sub3}
    \end{subfigure}
    \hfill
    \begin{subfigure}{0.24\textwidth}
        \centering
        \includegraphics[width=\textwidth]{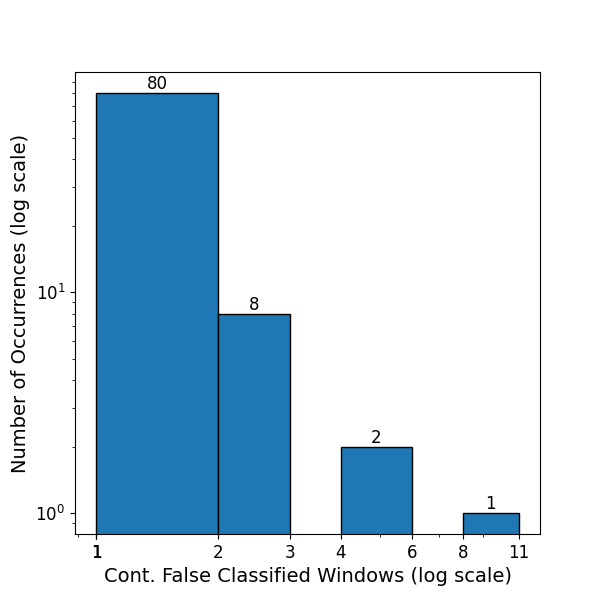}
        \caption{MM-FIT}
        \label{fig:sub4}
    \end{subfigure}
    
    \hfill
    \begin{subfigure}{0.24\textwidth}
        \centering
        \includegraphics[width=\textwidth]{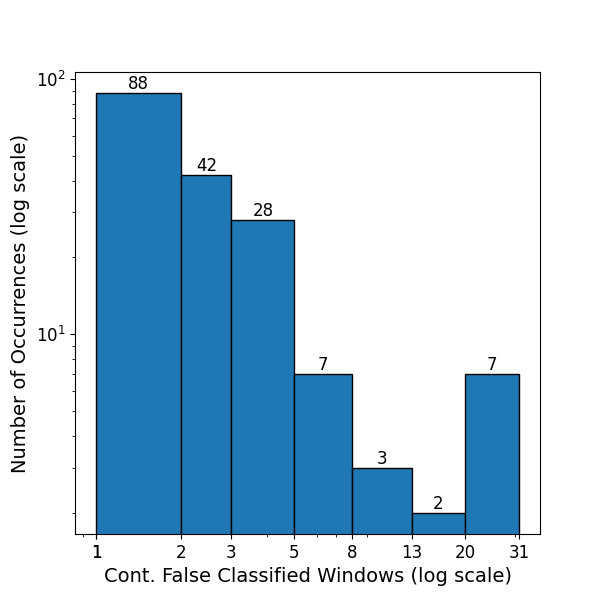}
        \caption{MHEALTH}
        \label{fig:sub5}
    \end{subfigure}
    \hfill
    \begin{subfigure}{0.24\textwidth}
        \centering
        \includegraphics[width=\textwidth]{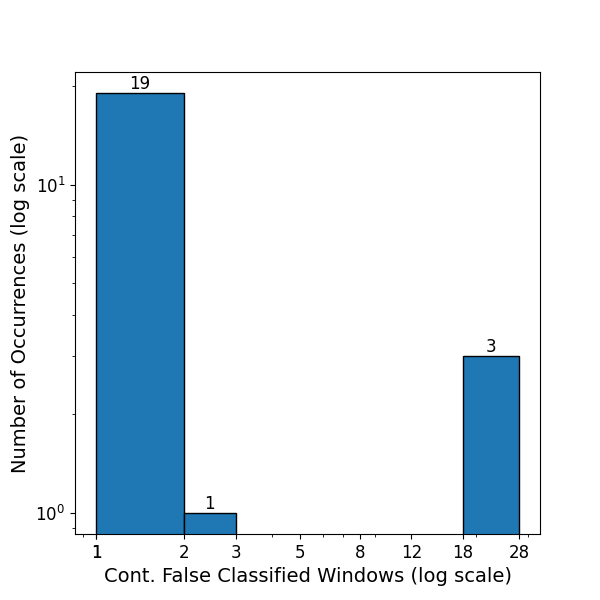}
        \caption{MotionSense}
        \label{fig:sub6}
    \end{subfigure}
    \hfill
    \begin{subfigure}{0.24\textwidth}
        \centering
        \includegraphics[width=\textwidth]{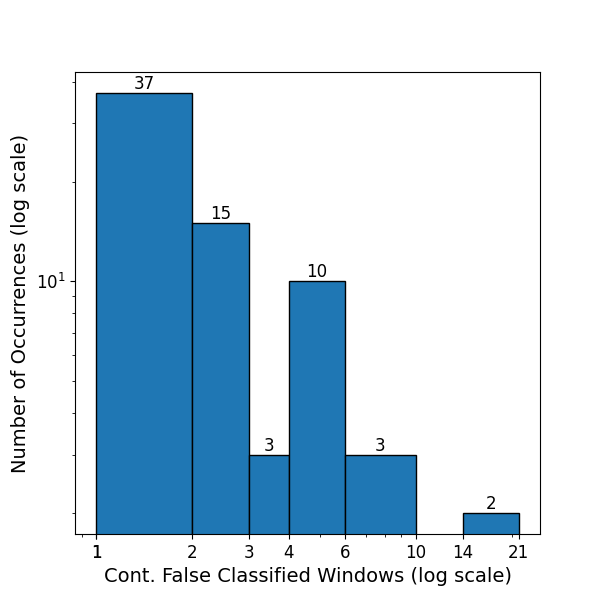}
        \caption{WISDM}
        \label{fig:sub7}
    \end{subfigure}
    \hfill
    
    \caption{Histograms (log-scale) for each dataset, representing the occurrence of continuous false classified windows.}
    \Description{}
    \label{fig:histogram}
\end{figure}

Next to inspecting the confused classes and their distribution, the number of continuous false classified windows offers insight into the contribution of each section to the overall confusion.
For each dataset, we calculated the continuity of wrong classified windows and visualized them in the form of histograms as shown in \cref{fig:histogram}.
Since either the number of occurrences on the y-axis as well as the duration of false classified windows on the x-axis contains a wide range across each bin, we decided to set both axes to a logarithmic scale to make them more meaningful.
Across all six datasets, trends can be derived towards higher occurrences of false classification with decreasing duration across the windows.
However, the findings can be divided into two parts, the commonly increased number of short-duration occurrences and a less prominent amount of long durations.

Starting with PAMAP2 in \cref{fig:sub1}, there are multiple sections in the dataset with long durations of false classification of more than 100 windows length.
Arguably, these parts may obtain some major issues within the data originating either from the annotation design or directly from the sensor data.
Similar findings can be drawn from the other dataset histograms with less exaggerating duration.
Nonetheless, focusing on MHealth and MotionSense in \cref{fig:sub4} and \cref{fig:sub5}, even though the durations of the false classified sections are smaller than for PAMAP2, the percentile contribution still requires proper investigation.
For the high occurrence of short false classification of windows, commonly in the range of one to 10 windows duration, the trained models may be uncertain about the current windows of sensor data, for instance, due to similar activity classes or disturbances from experiment conduction.
To identify the cause of such false classification sections, an in-depth visualization of appropriate windows, selected based on the findings from this and the previous section, will be provided in the following to conclude our experiments.

\section{Granular Inspection}
\label{sec:visual}
So far, we only stressed the the ambiguities within the dataset and their occurrence patterns and characteristics.
However, there needs to be an identification of the source for the false classification of windows.
Since there is commonly no video material of the experiment conduction available, mainly due to the data protection and anonymity of the participants, we established a visual inspection of the raw sensor signals to examine the IFC causes.
For each dataset and leave-one-out session, we created a plot that hosts the selected streams of sensor data underlaid with the IFC to visually distinguish the true from false classifications through green and red colors.
The x-axis represents the number of classified windows whereas one entry represents the sensor averaged sensor data of the full window.
Through this process, we obtain a condensed representation of the full experiment conduction, to investigate the correlation of sensor data to class prediction.

\begin{figure}[ht]
    \centering
    \includegraphics[width=\textwidth]{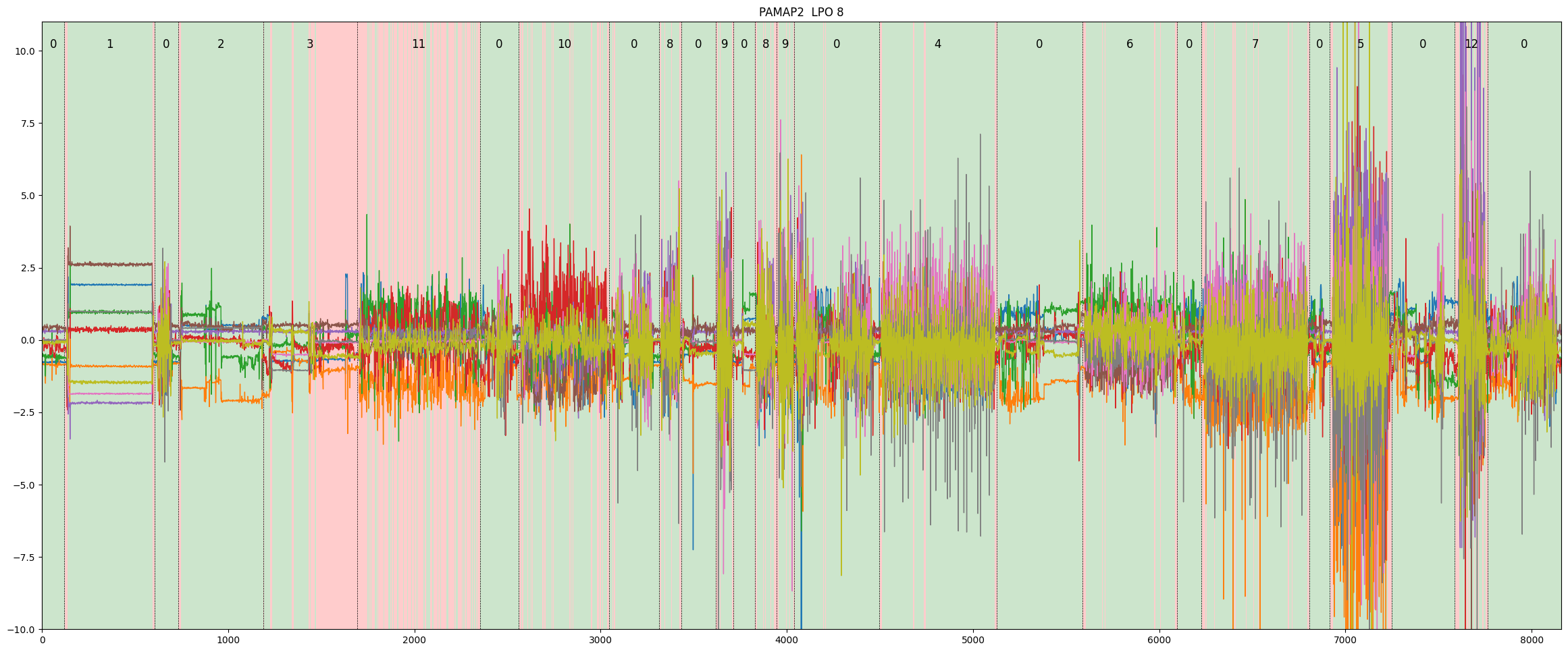}
    \caption{Exemplary view (PAMAP2 of Participant 1) visualizing the acceleration data of the three IMUs underlaid with correct and false classified activity class windows in green and red.}
    \label{fig:window_view}
\end{figure}

As shown in \cref{fig:window_view}, the exemplary view of PAMAP2 of Participant 1 outlines the classification issues, with large sections of wrong classification as well as short, alternating sections of confusion.
We additionally added vertical lines and the activity class IDs to outline the transitions between activities.
The corresponding labels for each id can be derived from \cref{sec:statistics}.
The sensor data of this figure shows three IMU acceleration data of PAMAP2. 
Since acceleration data covers the fine-grained movements of posture change in activities, it is more suitable than the gyroscope or magnetometer data for our condensed visualization
The full set of visualizations across all datasets was added in \cref{visual_compressed} as a reference.
Since they lack clear resolution of sensor data, we recommend them only as an initial check to obtain a decent overview of classification quality.

We selected sections of false classifications within the condensed representations through visual inspection to further investigate the source of misclassification.
The following sections categorize the IFC and present examples for each of them.
Each of the listed visualizations represents the highest resolution rate by stretching out the windows from the condensed plots into the granularity of data samples to thoroughly investigate the behavior of each sensor stream.
This is the fundamental reason why we are not able to visualize all ambiguous cases in such detail.
Within section \cref{sec:solution}, we sum up all the inspected sections and present a suitable adaption for each dataset

\subsection{Transition of Activity Classes}

\begin{figure}[!t]
    \centering
    \includegraphics[width=\textwidth]{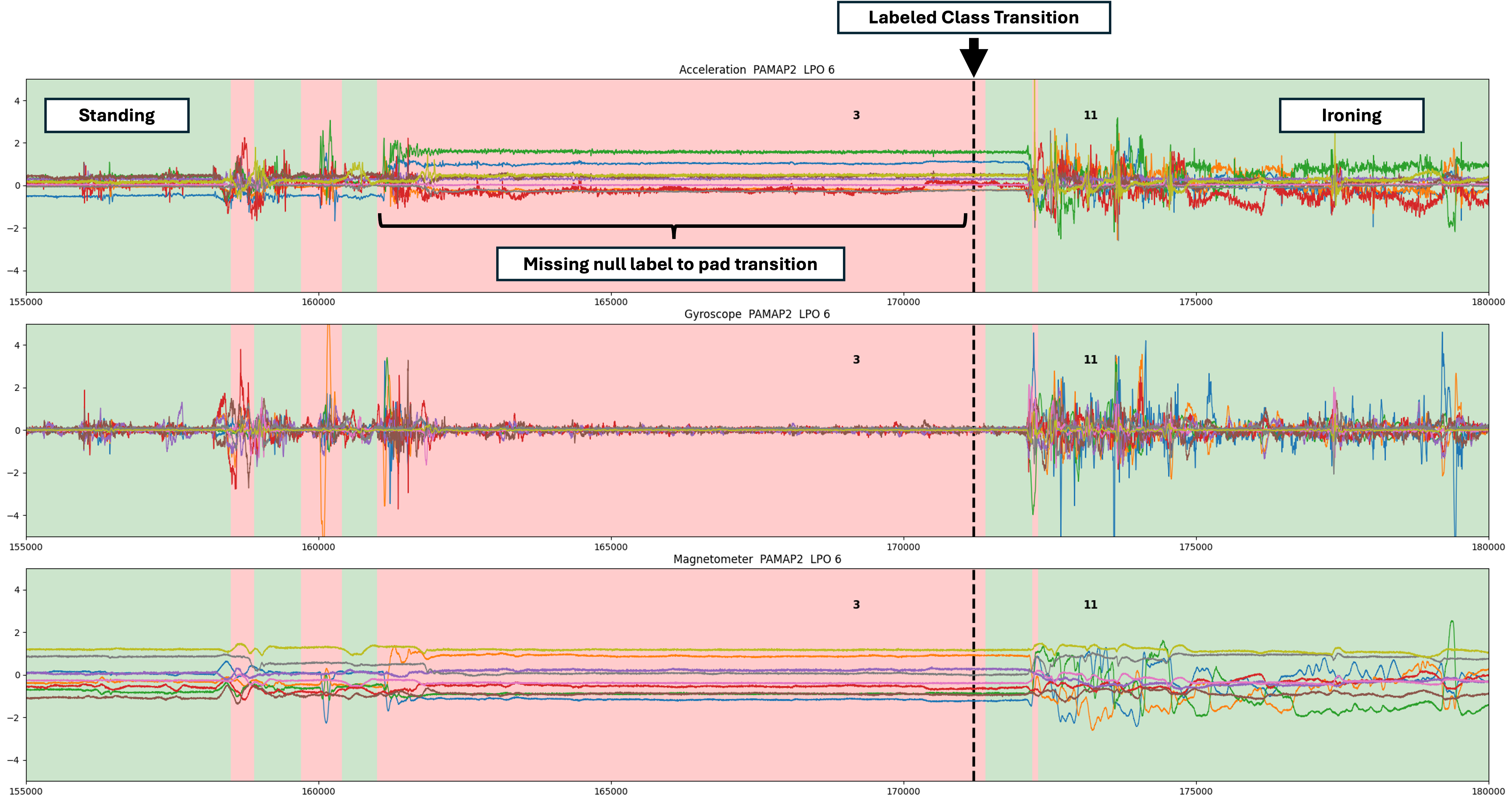}
    \caption{Confusion between \textit{Standing} and \textit{Ironing} in PAMAP2; the IFC shows false classified activities during transition}
    \Description{}
    \label{fig:data_pamap2_null}
\end{figure}

While comparing the datasets, we recognized variations in the labeling design of activity classes, especially when it comes to the granularity during transitions of classes.
As shown in \cref{fig:data_pamap2_null}, we selected the participant 6 recording of PAMAP2 with a focus on the transition between class 3 (\textit{standing still}) and 11 (\textit{ironing}).
Within the false classified classes, there is a clear change in acceleration data due to movements.
Even though the long section obtains an almost consistent signal, the values differ from those before.
In this case, it may be assumed the participant already prepared for the ironing activity.
Compared to other properly classified transitions, which are commonly connected via short \textit{null} class duration in between, the introduced movements labeled within the \textit{standing} class resulted in misclassifications of the missing \textit{null} class.

\subsection{Data Quality Issues}

\begin{figure}[!t]
    \centering
    \includegraphics[width=\textwidth]{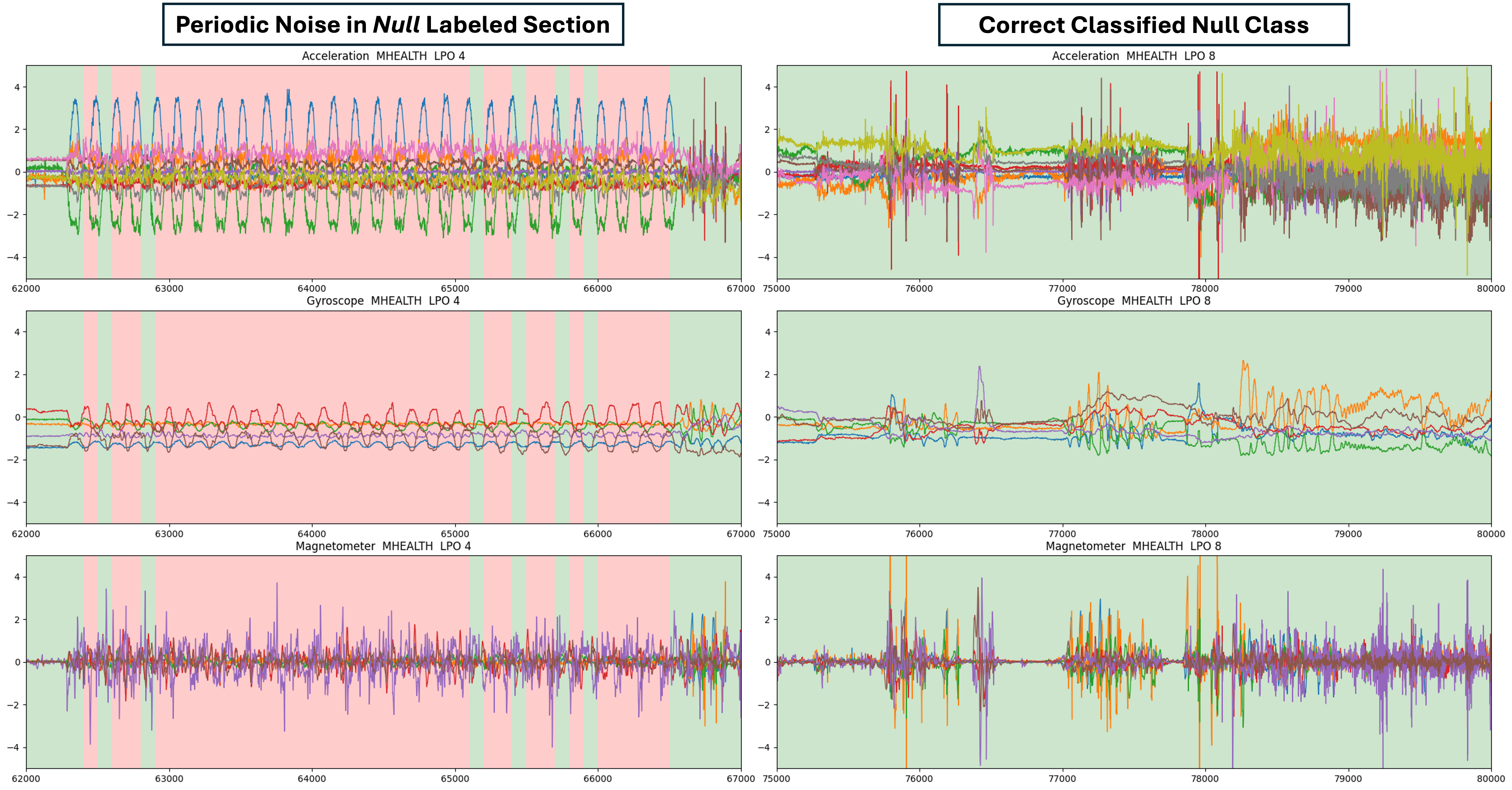}
    \caption{Periodic noise resulting in false classification in MHealth dataset (\textit{Null} labeled class is predicted as \textit{Cycling}; right represents correct classified \textit{null} class. }
    \Description{}
    \label{fig:data_mhealth_null}
\end{figure}

Symptomatic patterns in the recording quality of IMU sensors, especially when introduced through weak sensor attachment or hardware glitches, usually manifest themselves in false classifications due to misleading machine learning model predictions.
Within the MHealth dataset, we recognized some unusual patterns of uniform and highly frequented movements.
Due to the lack of synchronized video footage to support our findings, we are unable to locate the source of such a pattern.
As shown in \cref{fig:data_mhealth_null}, the left column states the selected section of Participant 4 from the MHealth dataset divided into the IMU modalities of the three sensors.
The section was labeled as \textit{null}, which, however, was confused with cycling due to the periodic signal. 
On the right side, we added a \textit{null} labeled section from participant 8, being classified correctly even though various changes in the signal can be recognized.
Although the null class of MHealth commonly includes movements and different postures, it however cannot handle such tremendous interference.

\subsection{Activity Uncertainty}
\begin{figure}[!t]
    \centering
    \includegraphics[width=\textwidth]{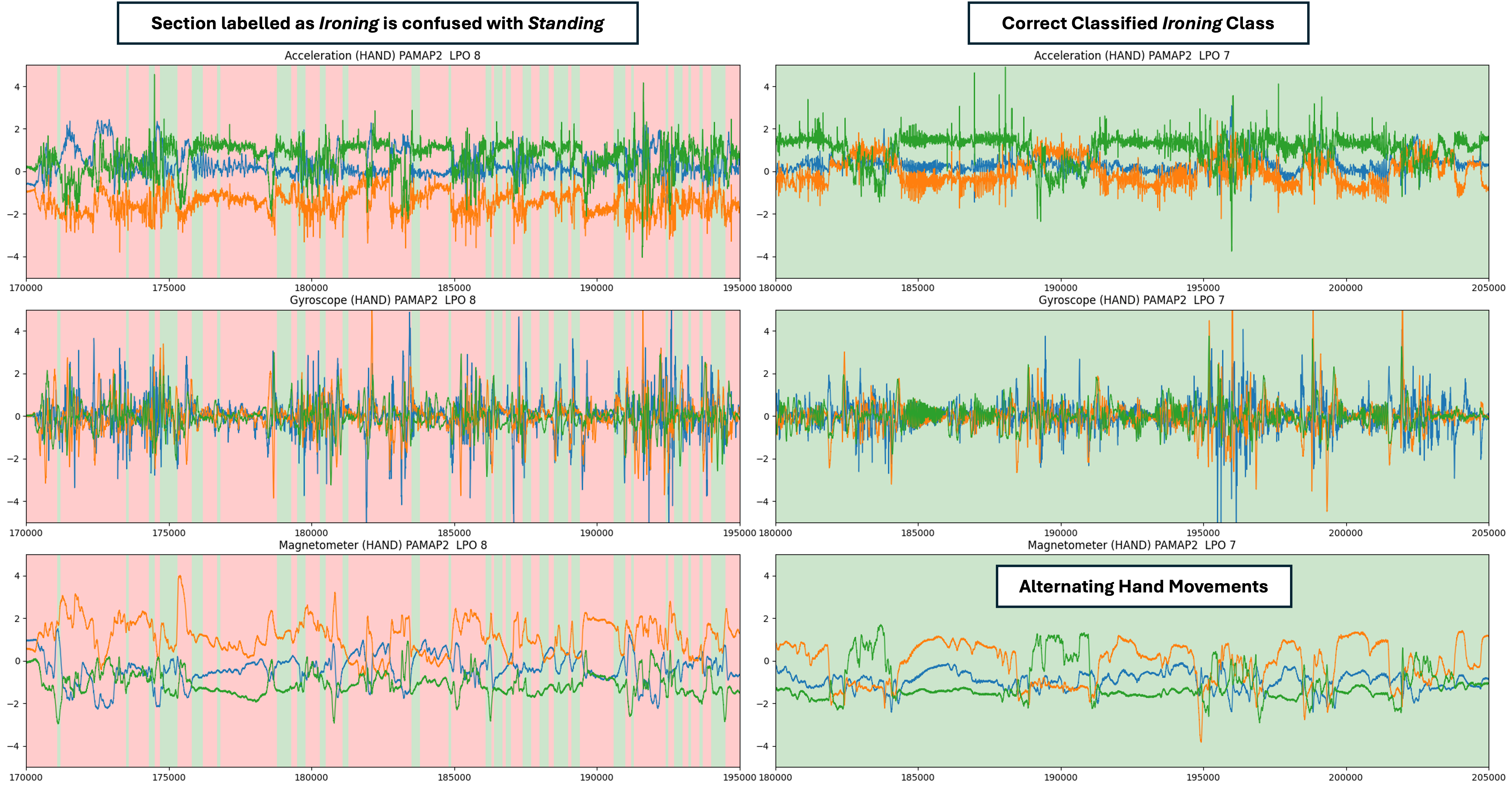}
    \caption{Ironing Activity of PAMAP2 compared between participants 8 (left) and 7 (right). The Magnetometer of Participant 7 shows clear alternating patterns from moving the iron whereas Participant 8 is confused with Standing.}
    \Description{}
    \label{fig:data_pamap2_ironing}
\end{figure}

The commonality of activities in the specific movements has always been a major problem in activity recognition to distinguish them properly. 
Although the tasks are completely different from a natural and task-orientated point of view, the execution and especially the recognized movements can have significant overlaps.
The confusion of \textit{standing} and \textit{ironing}, as already highlighted in \cref{sec:statistics} for the PAMAP2 dataset provides a great example of activity uncertainty and the importance of proper experiment conduction.
We plotted the two \textit{ironing} activities of Participant 8 (left) and Participant 7 (right) in \cref{fig:data_pamap2_ironing}, especially the hand IMU sensor since this sensor contains the most information on moving the iron as opposed to the \textit{standing} activity.
Comparing the visualizations, especially the magnetometer between the participants, the left sensor stream does not contain any useful information, leading to confusion with \textit{standing}.
On the opposite, the right magnetometer signal 
Due to the various execution methods in ironing, especially hand movement, recognition suffers if non-stringent movement patterns are executed.
In this example, the hand sensor of Participant 8 was unable to gather any valuable information to thoroughly guide the classification.

\section{Proposed Solution}
\label{sec:solution}

After inspecting the most frequently occurring challenges in HAR datasets, mainly due to the three particularities from the previous section, we now propose a solution to handle the calculated IFC.
We do not aim to rework the whole dataset, for instance by relabelling the activity classes, as this would change the entire nature of the data set and comparison with previous works and the traditional dataset would be impossible.
Rather, we introduce a mask, which a potential user can overlay onto his dataset and predictions in order to act as a filter for handling the IFC sections.

We defined a trinary mask to mark each section of the dataset with one of three possible categories as follows:
\begin{enumerate}[start=0]
    \item \textbf{clean}, issue free data, confidently classified correctly across our experiments

    \item \textbf{minor}, false classifications due to model uncertainty in experiments

    \item \textbf{major}, strong confidence towards false classifications throughout our experiments
\end{enumerate}

To utilize the results of our experiments toward the correct categorization of our trinary mask, we analyzed the prediction probabilities of each window as a basis to estimate the uncertainty of the prediction.
For each set of predicted class probabilities that is located within the IFC, we sort them from highest to lowest probability and calculate the distance to the next weaker probability.
If the largest distance can be found between the first two probabilities, it is an indicator of strong confidence in the prediction.
However, since this prediction is confidently falsely classified, we categorize it as major.
If the largest distance is located between probabilities from the lower order, we categorize it as minor since multiple classes contribute to the model prediction uncertainty.
For those, we assume that there still exist potential chances of correct classification beyond our experiments.

In \cref{tab:mask}, we show the distribution of the three defined labels for each dataset.
When comparing the selected categories and their distribution with the duration of false classified sections in \cref{sec:histograms}, the histograms show a clear correlation between the duration of IFC and the mask categorization.
We plausibly checked the automated categorization process through manual, visual inspection as introduced in \cref{sec:visual}, which demonstrated a trend of major labels commonly located in the longer sections of continuous false classification and minor labels located within the shorter sections.
This proves our assumption of uncertain and short false predicted sections due to data obscurities as opposed to longer, confident false predictions due to critical issues affecting proper classification.

\begin{table}[!t]
\footnotesize
  \centering
  \caption{The distribution of the proposed trinary mask for each dataset.}
  \begin{tabular}{cccc}
    \toprule
     & \textbf{Clean (0)} (\%)& \textbf{Minor (1)} (\%)& \textbf{Major (2)} (\%) \\
    \midrule
    \textbf{PAMAP2} & 88.26 & 8.16 & 3.58 \\  
    \textbf{Oppo-Locomotion} & 94.68 & 3.88 & 1.43 \\  
     \textbf{Oppo-Gesture}& 87.94 & 9.58 & 2.48 \\  
    \textbf{MM-FIT} & 99.56 & 0.36 & 0.08\\  
    \textbf{MHEALTH} & 95.42 & 3.17 & 1.41\\  
    \textbf{MotionSense} & 99.54 & 0.15 & 0.31\\  
    \textbf{WISDM} & 98.39& 0.61 & 1.00\\  
    \bottomrule
  \end{tabular}
  \label{tab:mask}
  
\end{table}

For our trinary IFC mask, we envision two application scenarios: First to mask out the predictions of the traditional training regime to verify and quantize the overlap between the predictions and our IFC mask; Secondly, directly apply the mask within the machine learning training process by for instance masking out major labels or by generating awareness about demanding sections through integration of IFC mask labels into the loss function.

The full mask providing the sequences of labels for each dataset will be publicly available.

\section{Discussion}
\subsection{Guideline for HAR Data-Acquisition}

Precise sensor data acquisition and accurate labeling are indispensable for proper HAR.
Based on our findings from the previous sections, we present a guideline for enhanced HAR experiments in the following:
\begin{itemize}
    \item Definition of Objectives: Define clear goals and objectives of the HAR system to ensure that the activities targeted for recognition are aligned with the intended application, facilitating human-natural data collection and model development. The collection of labels should be mutually exclusive from each other if merged in a single track to reduced label ambiguity. Or independent tracks of labels like the practices in Opportunity can be considered.
    \item Appropriate Sensor Selection and Placement: Select sensors that are suitable for the intended activities, especially so that the sampling rate and accuracy match the activities. Additionally, ensure they are securely placed, and calibrated on the body and obtain data in a synchronized manner to capture the activities properly.
    \item Participant selection: Consider age, gender, body size, and activity levels when selecting participants to ensure a diverse and representative dataset to reflect the population exposed to HAR.
    \item Experiment Environment: Select an environment that replicates real-world conditions, controlling factors like external disturbances outside of the laboratory setup. Next to the hAR sensing equipment, ensure proper secondary recording devices like visual or audio recorders to annotate the data.
    \item Experiment Conduction: Follow a standardized protocol for data collection, providing clear but open-minded instructions to participants to ensure consistency in the execution of activities and rest periods. Moreover, the execution has to feel natural to the participants so the experiment recording is unbiased from a staged environment.
    \item Data Annotation: Annotate collected data with ground truth labels indicating the type and timing of activities performed. Depending on the activities, fine-granulated labeling is inevitable to generate a qualitative dataset.
     
\end{itemize}
\subsection{Limitations}

Our approach aimed to investigate each dataset in a meaningful way to remove any influences coming from certain model architectures.
Therefore, we combined the false negative classes from each training procedure to build a common basis for our investigations.
We attempted to strike a balance within the wide range of available architectures to carve out the dataset ambiguities and challenges solely.
However, we couldn't establish a clear measure of our set of selected machine learning models sufficiently representing an appropriate intersection.
It remains a question for future work if out-performing model architectures would contribute to our inspection, whether they advantageously extract the required information from the minor or major labeled mask, or if they can only recognize the problems better and handle them correctly.

Additionally, splitting the time-series data into windows, even though it currently represents state-of-the-art practice, may lead to differences in classification performance due to the split.
Depending on the activity and according to pattern duration, the time-series data may be split unfortunately.
Especially for datasets containing related activities of similar alternating duration, splitting data at an inappropriate section can be dragged through the whole dataset. 
An extended validation based on the proposed trinary mask will be incorporated in future works to investigate the current limitations on subjected model architectures, variations in window size, and the effects of varying the stride.

Due to the unavailability of synchronized video material for most of the datasets, except for the works of Opportunity++, we cannot inspect what exactly happened throughout each recording session.
However, we aim to investigate further into the source identification of false classifications.

\section{Conclusion}
To conclude, motivated by the dialectic questions about the mainstream HAR research approach in \cref{fig:intro}, our granular examination of HAR benchmarking datasets first have identified and quantified segments in six popular benchmark datasets which cannot be correctly classified by any existing ML models, which we denote as the intersect of false classifications (IFC).
Detailed analysis of the IFC showcases the crucial challenges of this field with in-depth considerations towards the sources of the false classifications.
One of the primary observations from our analysis is the variability in labeling practice which could inherently introduce ambiguity, particularly evident during transitions between activity classes, especially with increased transitions of null classes incorporated in the dataset.
Additionally, labeling design can introduce inherent ambiguity for single-output classification tasks, as activities with overlapping movement patterns or shared similarities in execution presents another challenge.

To address those challenges and provide avenue for possible solutions, we proposed a trinary mask as an additional dataset annotation layer to filter and categorize the time-series data into three categories. 
While the trinary mask outlines a promising solution to tackle the HAR challenges, future research efforts should aim to seek for methods incorporating the trinary mask.
Our work reveals the granular nuances and thus we encourage future ML research on HAR to look beyond the statistics and provide more granular discussion.
We also hope to inspire data collection efforts with practices that improves the auditability of HAR datasets.
\begin{acks}

\end{acks}

\bibliographystyle{ACM-Reference-Format}
\bibliography{references}

\newpage
\appendix

\section{Model Architecture Details}
\label{model_details}
The following summary explains the utilized models from our experiments in more detail:

Our Convolutional Neural Network (\textbf{CNN}) architecture comprises convolutional blocks for feature extraction followed by a multi-layer perceptron (MLP) block for classification. 
Specifically, there are three convolutional blocks with kernel sizes of 24, 16, and 8 respectively, and feature sizes of 32, 64, and 128 for each block respectively. 
Between each block, a ReLU activation layer and dropout with a rate of 0.1 are applied for regularization. 
Max pooling is performed on the features before the dense layer. 
The classifier consists of two dense layers, with the first outputting a feature size of 1024, and the last one predicting classes.

The utilized \textbf{GRU} consists of two consecutive Gated Recurrent Unit (GRU) blocks, each comprising two hidden layers with 128 units. 
Following the GRU blocks, there is a single dense layer serving as the classifier, with the target dataset labels. 
Before the dense layer, batch normalization and dropout techniques are applied. 
Specifically, dropout with a rate of 0.5 is employed to regularize the network and prevent overfitting.

The \textbf{LSTM} network architecture is composed of two consecutive LSTM blocks, each consisting of two hidden layers with 128 units in each layer. 
Subsequently, a single dense layer acts as the classifier with the target dataset labels. 
Prior to the dense layer, we apply batch normalization and dropout techniques. We specifically utilize dropout with a rate of 0.5 to regularize the network and mitigate overfitting. Our LSTM and GRU have similar settings.

Our \textbf{ConvLSTM} network architecture consists of a feature extractor followed by a classifier. 
The feature extractor comprises three convolutional blocks, each consisting of convolutional layers with a kernel size of 5 and 64 features, followed by a ReLU activation function. 
Following the convolutional blocks, there is a single LSTM layer with 128 units. 
Prior to the classifier, dropout with a rate of 0.5 is applied for regularization. 
The class probabilities are predicted by a dense layer.

The Transformer-based network architecture and framework for HAR (\textbf{TinyHAR}) features four individual convolutional layers, followed by a single self-attention block with one head, applied channel-wise. 
Dropout is applied to the features to enhance regularization. 
The vectorized representations are then processed through a dense layer with the ReLU activation function. 
To capture temporal dependencies, the network incorporates Long Short-Term Memory (LSTM), which outputs temporally weighted representations. Finally, a dense layer utilizes these representations for predicting the classes.

Contrastive Predictive Coding (\textbf{CPC}) \cite{cpc:har,oord2018representation:cpc} is a model enhancement method through self-supervised pre-training of the feature encoder through a task of recurrently predicting the future latent features. 
CPC was selected among many self-supervised techniques both for brevity and its only requirement of continuous sensor data making it easily applicable to all of the datasets. 
The encoder was unlocked in the downstream classification task on the dataset, as it was shown to be more effective in \cite{fortes2023don}.  
CPC predicts multiple future time steps, leveraging long-term temporal patterns. 
This approach works well for activity recognition and semi-supervised learning, particularly in scenarios with multiple sensors.
The encoder was pretrained on the same dataset without annotation and the pre-trained weights were utilized for classification afterward. 
In our settings the networks were not frozen as in the original paper \cite{cpc:har}. 
    
\section{Window-Based Visual Inspection}
\label{visual_compressed}
The following graphics provide the window-based visualization for each dataset and cross-validation setup with time-synchronized sensor data underlaid with a red background for the false classified sections extracted from the IFC calculations.
In order to maintain interpretability, we only show the acceleration data of each sensor.

\begin{figure}[ht]
    \centering
    \includegraphics[width=\textwidth]{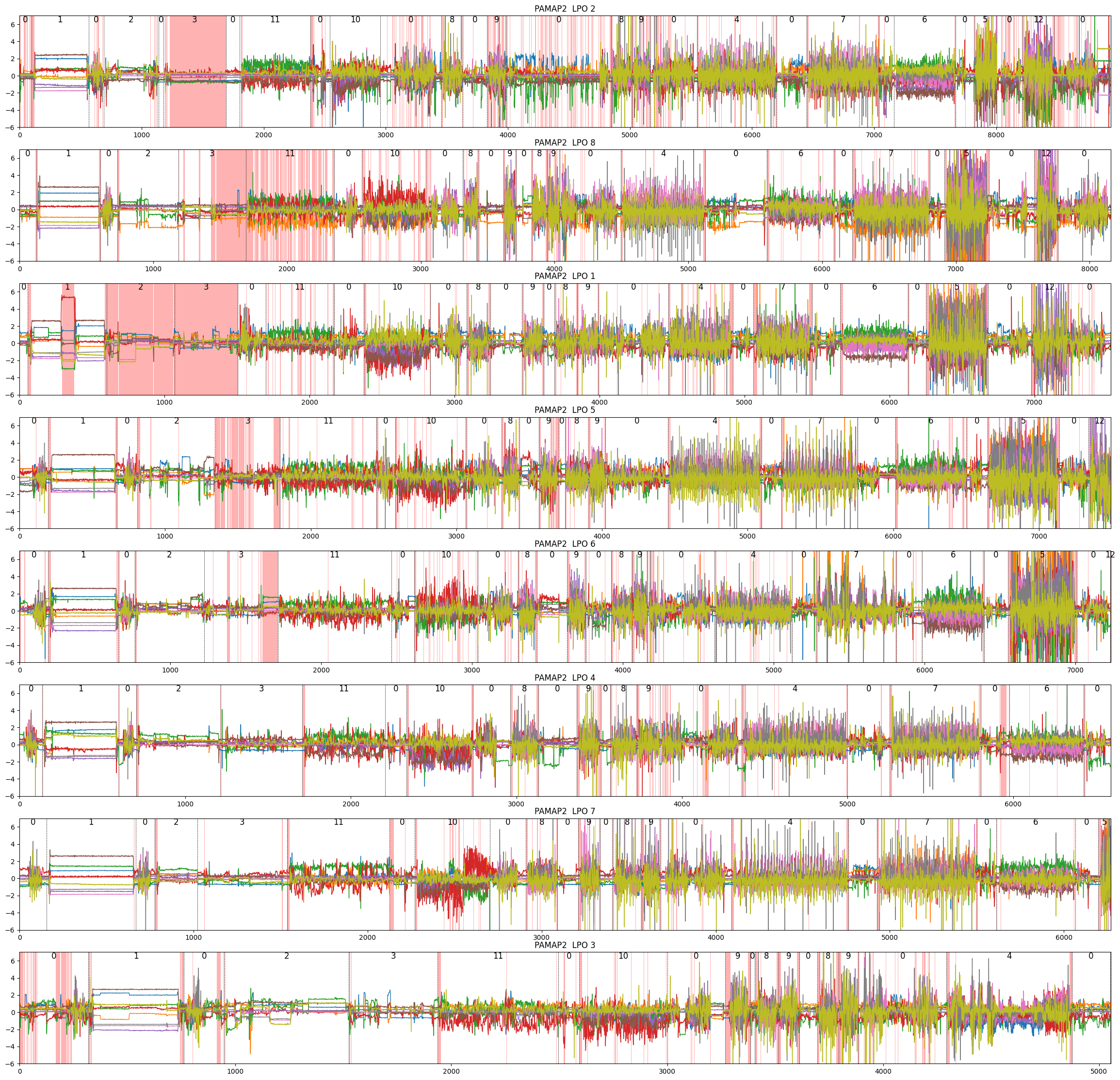}
    \caption{Sensor Data and False Classifications of PAMAP2}
    \Description{}
    \label{fig:pamap2_app}
\end{figure}
\begin{figure}[ht]
    \centering
    \includegraphics[width=\textwidth]{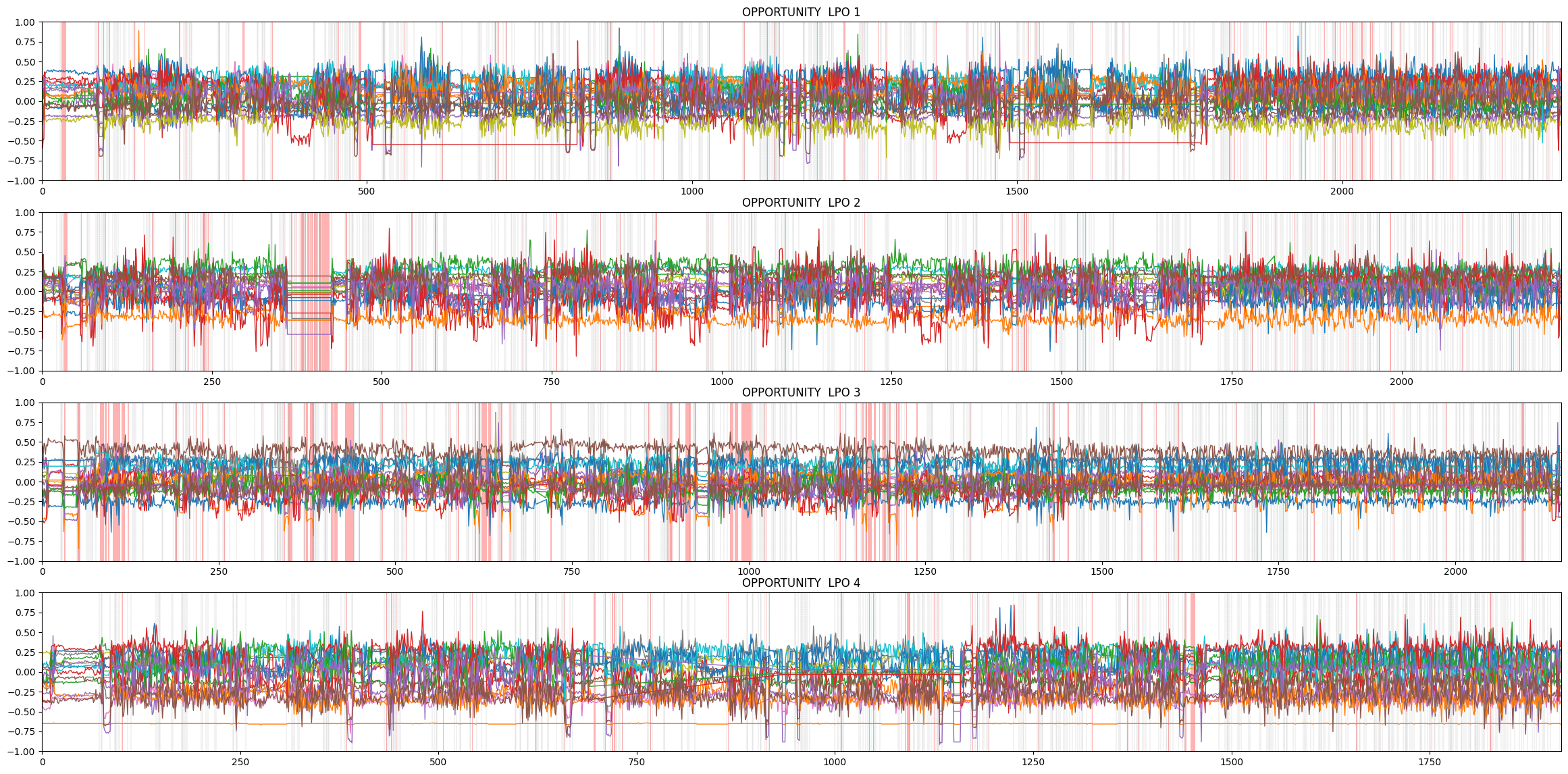}
    \caption{Sensor Data and False Classifications of OPPORTUNITY}
    \Description{}
    \label{fig:oppo_app}
\end{figure}
\begin{figure}[ht]
    \centering
    \includegraphics[width=\textwidth]{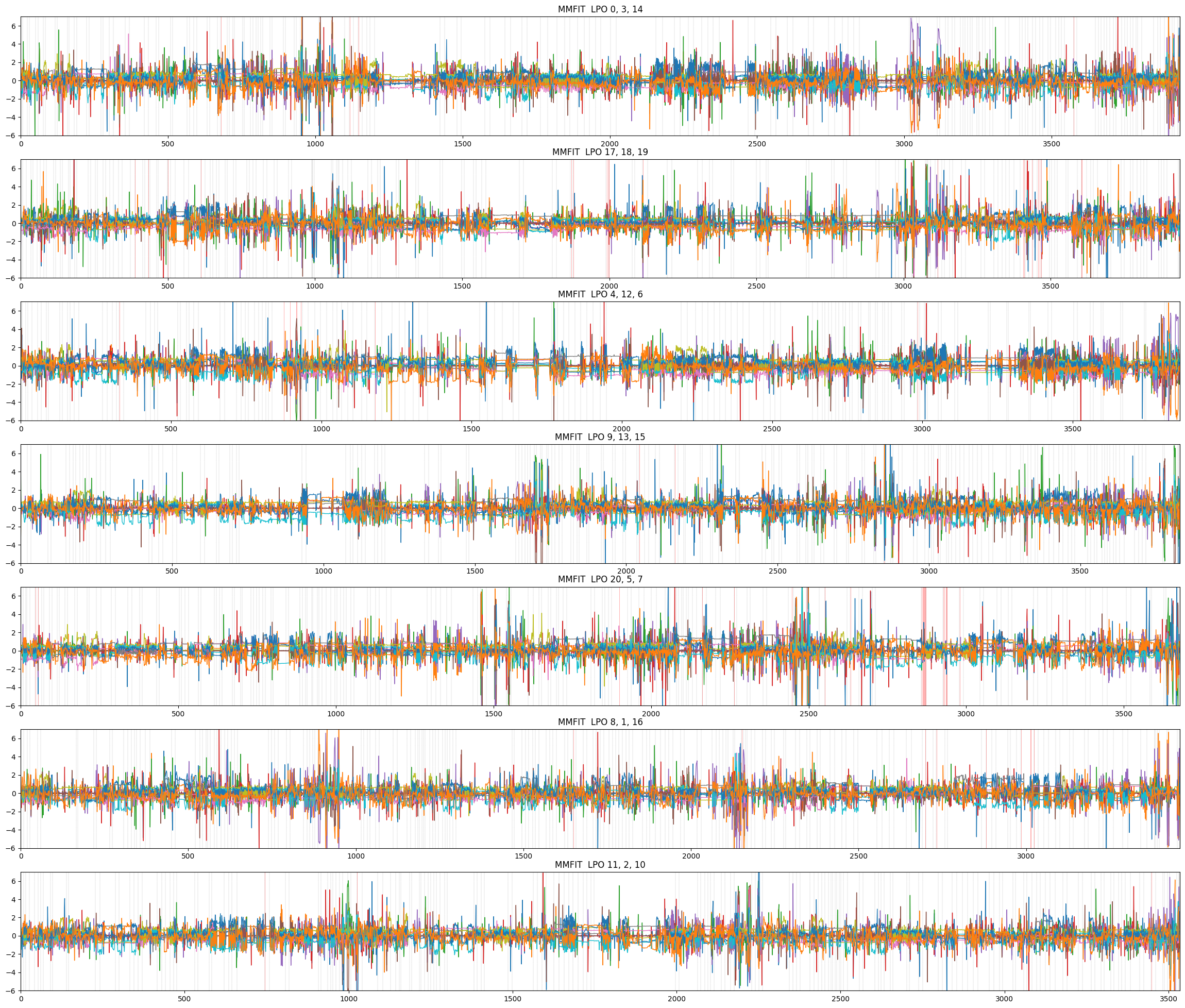}
    \caption{Sensor Data and False Classifications of MM-FIT}
    \Description{}
    \label{fig:mmfit_app}
\end{figure}
\begin{figure}[ht]
    \centering
    \includegraphics[width=\textwidth]{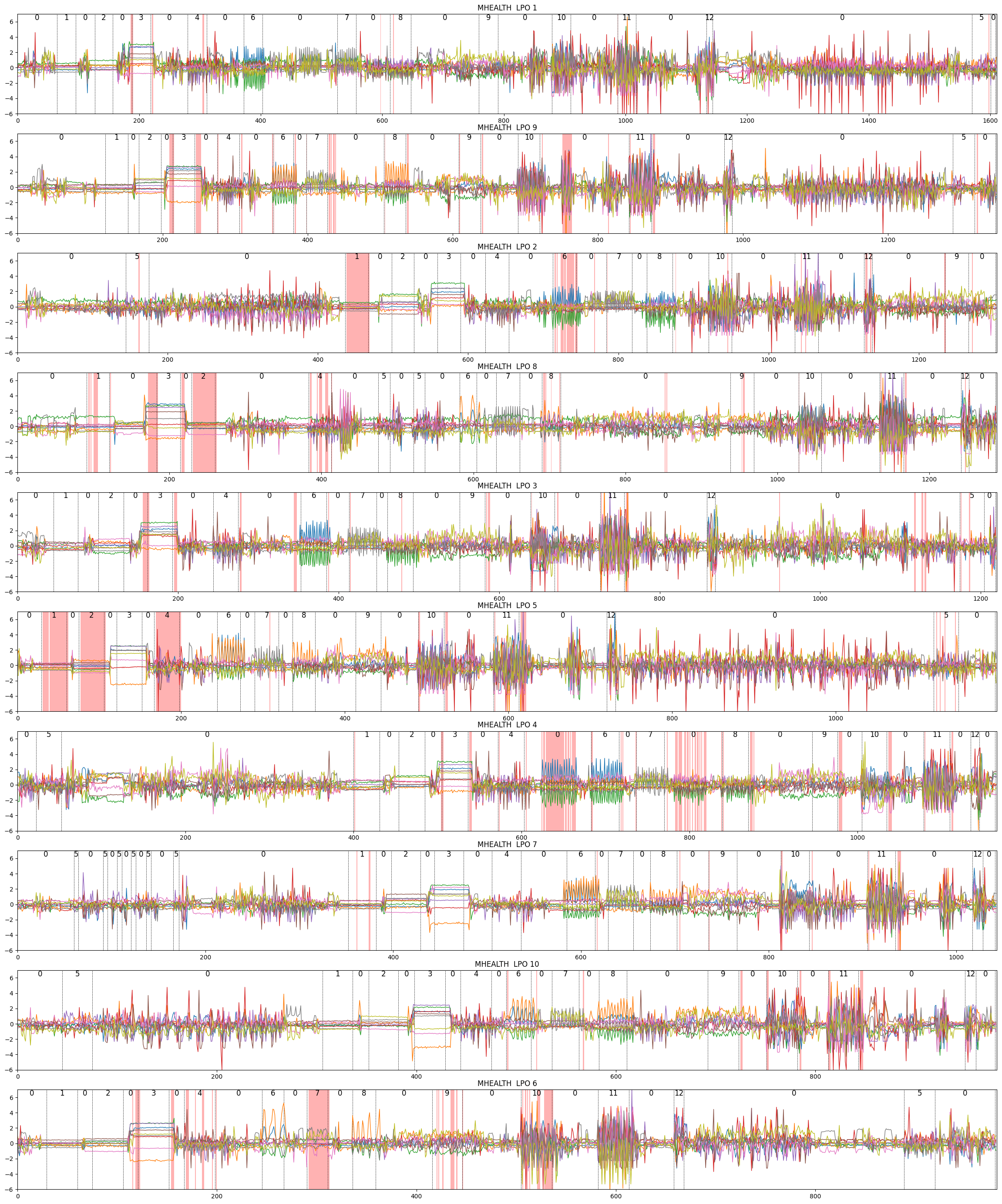}
    \caption{Sensor Data and False Classifications of MHEALTH}
    \Description{}
    \label{fig:mhealth_app}
\end{figure}
\begin{figure}[ht]
    \centering
    \includegraphics[width=\textwidth]{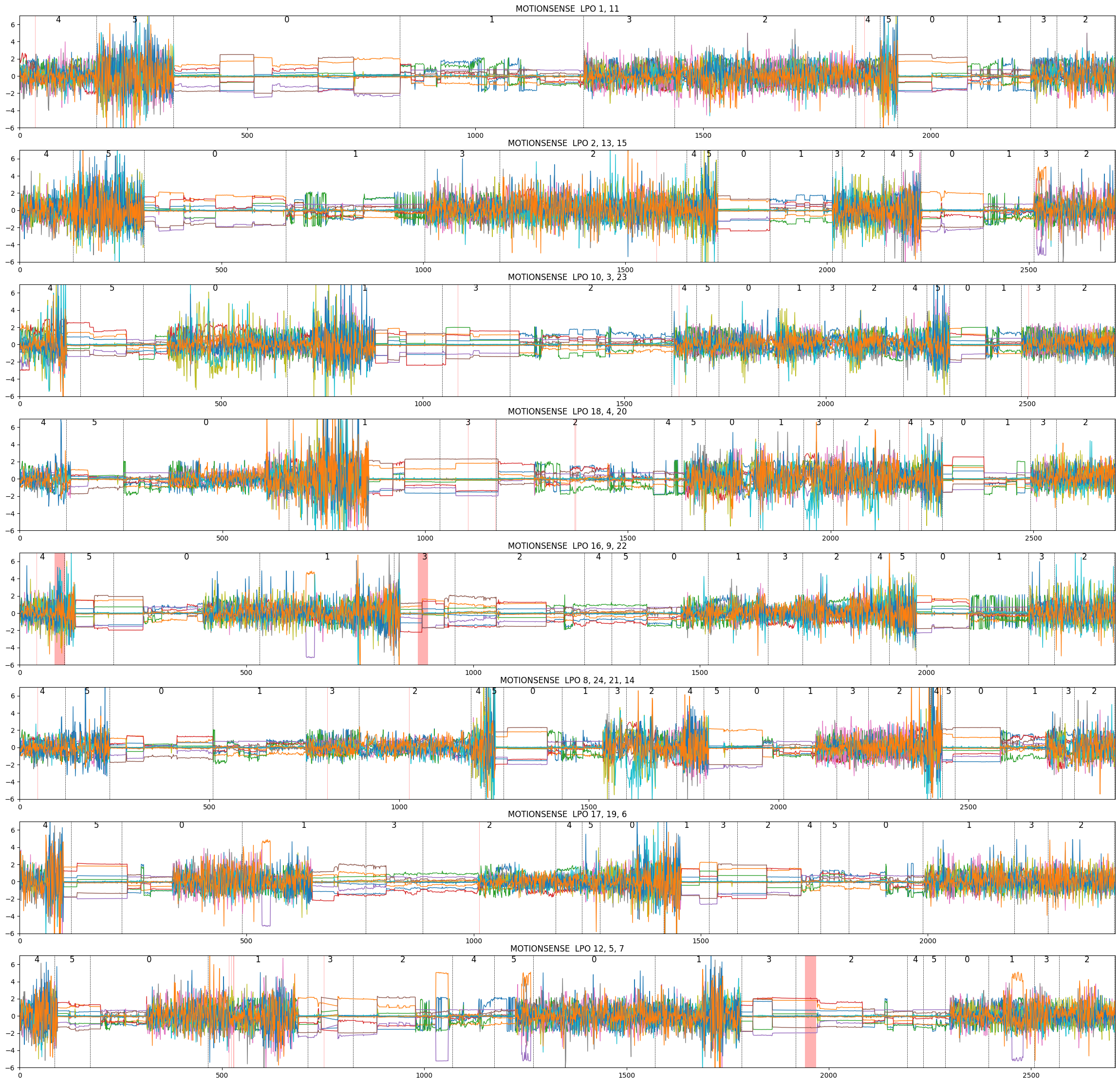}
    \caption{Sensor Data and False Classifications of MotionSense}
    \Description{}
    \label{fig:motion_app}
\end{figure}
\begin{figure}[ht]
    \centering
    \includegraphics[width=\textwidth]{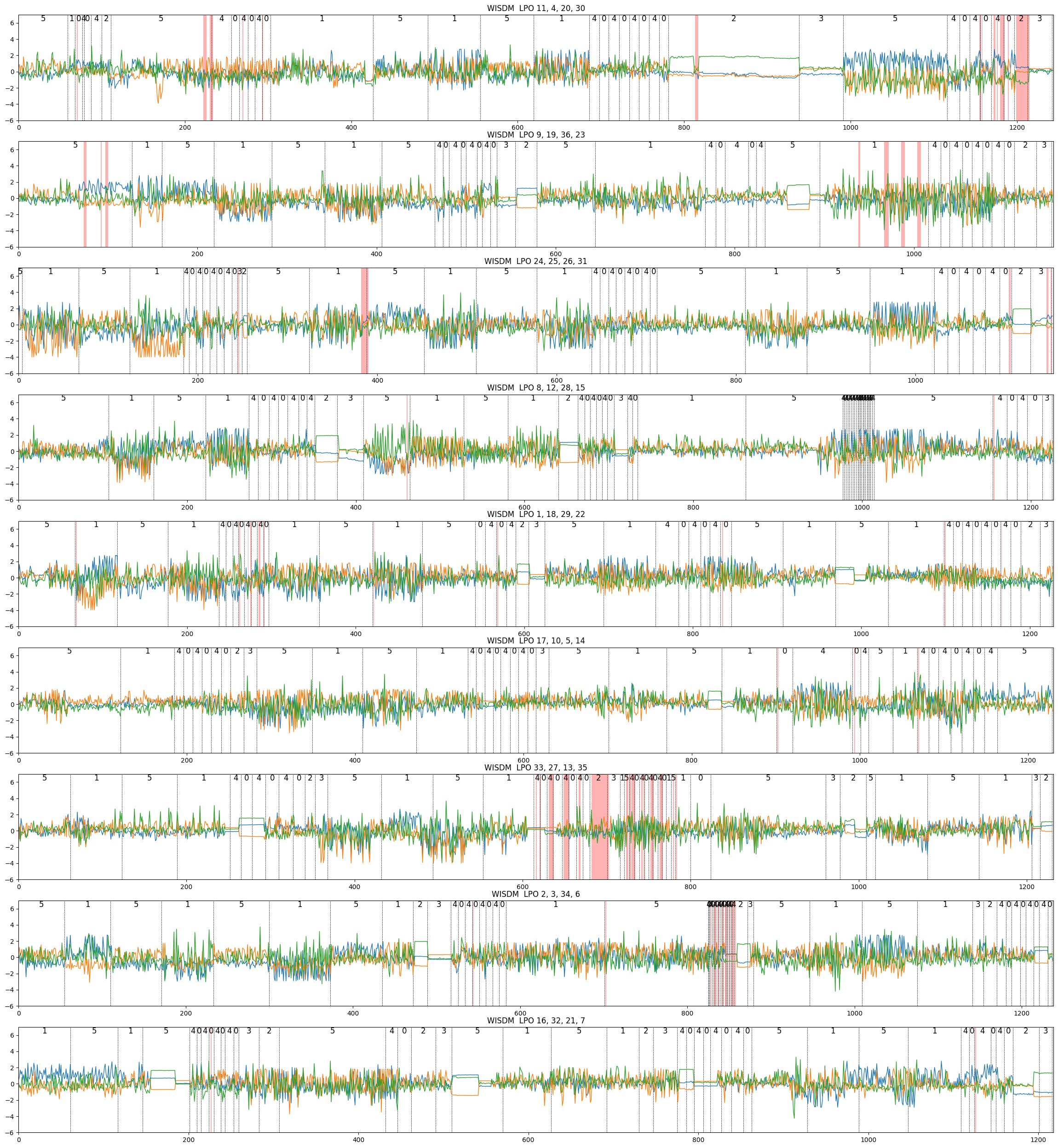}
    \caption{Sensor Data and False Classifications of WISDM}
    \Description{}
    \label{fig:wisdm_app}
\end{figure}

\end{document}